%% file: main.tex
\documentclass[10pt,twocolumn,letterpaper]{article}

\usepackage[pagenumbers]{cvpr} 


\usepackage[utf8]{inputenc} 
\usepackage[T1]{fontenc}    
\usepackage[pagebackref,breaklinks,colorlinks]{hyperref}       
\usepackage{url}            
\usepackage{booktabs}       
\usepackage{amsfonts}       
\usepackage{nicefrac}       
\usepackage{microtype}      
\usepackage{algorithm}
\usepackage{algpseudocode}
\usepackage{amsmath}
\usepackage{amssymb}
\usepackage{amsthm}
\usepackage{enumitem}
\usepackage{gensymb} 
\usepackage{multirow, makecell}
\usepackage{soul}
\usepackage{bbm}
\usepackage{subcaption}
\usepackage{xcolor, color, colortbl}
\usepackage{graphicx}
\usepackage{algorithmicx}
\usepackage{mathtools}
\usepackage{adjustbox}
\usepackage{import}
\usepackage{caption}
\usepackage{tabularx}
\usepackage{rotating}
\usepackage{xspace}
\usepackage{tikz}
\usetikzlibrary{spy,backgrounds}
\usepackage{arydshln}
\usepackage{cuted}

\usepackage[capitalize]{cleveref}

\crefname{section}{Sec.}{Secs.}
\Crefname{section}{Section}{Sections}
\Crefname{table}{Table}{Tables}
\crefname{table}{Tab.}{Tabs.}
\input{mycommand}

\newcommand{\methodname}{GST\xspace}
\newcommand{\longmethodname}{Gaussian Splatting Transformer}

\definecolor{cvprblue}{rgb}{0.21,0.49,0.74}

\def\paperID{41} 
\def\confName{CVPR}
\def\confYear{2025}

\title{GST: Precise 3D Human Body from a Single Image with Gaussian Splatting Transformers\\}

\author{
Lorenza Prospero\textsuperscript{1,2} \quad
Abdullah Hamdi\textsuperscript{2} \quad
Joao F. Henriques\textsuperscript{2} \quad
Christian Rupprecht\textsuperscript{2} \\
\normalsize \textsuperscript{1}The Podium Institute for Sports Medicine and Technology, University of Oxford \\
\normalsize \textsuperscript{2}Visual Geometry Group, University of Oxford \\
\small \tt{\{lorenza,abdullah,joao,chrisr\}@robots.ox.ac.uk}
}

\begin{document}
\twocolumn[{
\maketitle
\vspace{-0.5cm} 
\centering
  \begin{tikzpicture}[font=\footnotesize]
    \centering
\node (A) [anchor=west] at (0,0) {\includegraphics[width=0.95 \textwidth]{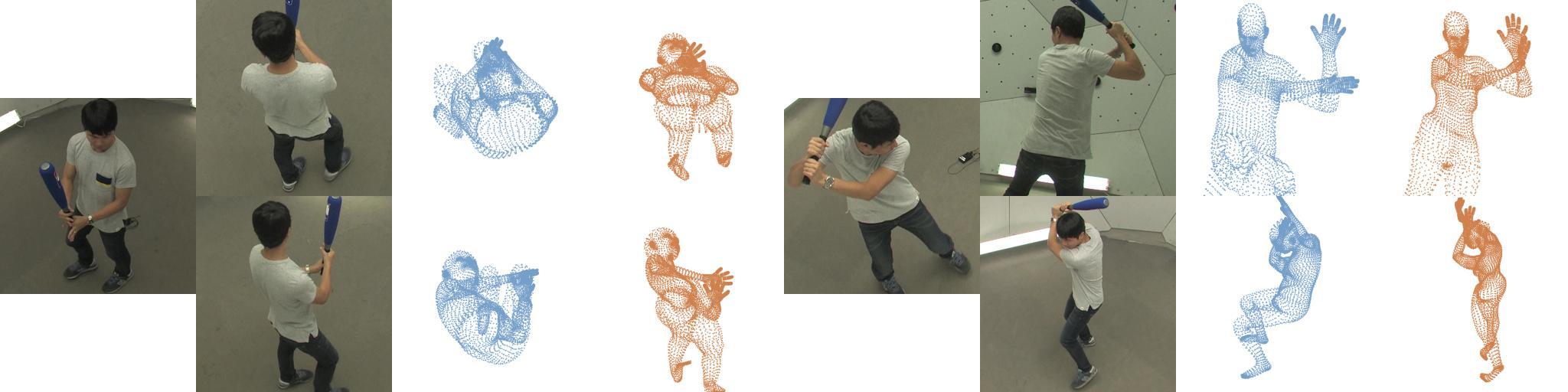}};
\node (B) [anchor=west] at ([xshift=-8.2cm, yshift=0.1cm] A.north) {Input Image};
\node (C) [anchor=west] at ([xshift=-6cm, yshift=0.1cm] A.north) {Other Views};
\node (D) [anchor=west] at ([xshift=-3.7cm, yshift=0.1cm] A.north) {HMR2 \cite{4dhumans}}; 
\node (E) [anchor=west] at ([xshift=-1.8cm, yshift=0.1cm] A.north) {\textbf{GST (ours)}};
\node (F) [anchor=west] at ([xshift=0.2cm, yshift=0.1cm] A.north) {Input Image};
\node (G) [anchor=west] at ([xshift=2.2cm, yshift=0.1cm] A.north) {Other Views};
\node (D) [anchor=west] at ([xshift=4.5cm, yshift=0.1cm] A.north) {HMR2 \cite{4dhumans}};
\node (E) [anchor=west] at ([xshift=6.5cm, yshift=0.1cm] A.north) {\textbf{GST (ours)}}; 
\end{tikzpicture}
\captionsetup{type=figure}
\captionof{figure}{\textbf{Example of human pose improvements using our method \methodname.} 3D human body results of our \methodname and SMPL predictions of HMR2 \cite{4dhumans} on a sports sequence from the CMU panoptic dome dataset \cite{cmu}.}
\label{fig:teaser}
\vspace{0.2cm} 
}]

\raggedbottom

\input{sections/abstract}
\input{sections/introduction}
\input{sections/relatedwork}

\input{sections/methodology}

\input{sections/experiments}
\input{sections/results}
\input{sections/ablation}

\input{sections/conclusion}


{
\small
\bibliographystyle{ieeenat_fullname}
\bibliography{main}
}

\clearpage
\begin{strip}
    \centering
    {\Large \textbf{GST: Precise 3D Human Body from a Single Image with Gaussian Splatting Transformers}} \\ \vspace{2em} \Large Supplementary Materials \vspace{2em}
\end{strip}

\appendix
\input{sections/supplement}

\end{document}

%% file: mycommand.tex
\usepackage{bm}

\newcommand{\minorsection}[1]{\noindent\textbf{#1.}}
\newcommand{\supp}{\textit{the Appendix}\xspace}

\newcommand{\REMOVAL}[1]{}

\definecolor{Highlight}{HTML}{39b54a}  %

\usepackage{amssymb}%
\usepackage{pifont}%
\newcommand{\cmark}{\color{red}{\ding{51}}}%
\newcommand{\xmark}{\color{black}{\ding{55}}}%

\definecolor{rblue}{RGB}{68, 114, 196}
\definecolor{rgreen}{RGB}{112, 173, 71}
\definecolor{rorange}{RGB}{237, 125, 49}

\newcommand{\cmmnt}[1]{}

\usepackage[normalem]{ulem}

\usepackage{xcolor}
\usepackage{colortbl}
\definecolor{turquoise}{cmyk}{0.65,0,0.1,0.3}
\definecolor{purple}{rgb}{0.65,0,0.65}
\definecolor{dark_green}{rgb}{0, 0.5, 0}
\definecolor{orange}{rgb}{0.8, 0.6, 0.2}
\definecolor{red}{rgb}{0.8, 0.2, 0.2}
\definecolor{blueish}{rgb}{0.0, 0.7, 1}
\definecolor{light_gray}{rgb}{0.7, 0.7, .7}
\definecolor{pink}{rgb}{1, 0, 1}
\definecolor{dark_red}{rgb}{0.5, 0, 0}

\definecolor{gold}{HTML}{EBC634}
\definecolor{silver}{HTML}{B4B4B4}
\definecolor{bronze}{HTML}{AD8A56}

\definecolor{light_silver}{HTML}{EBEBEC}
\definecolor{light_gold}{HTML}{FFF080}
\definecolor{light_bronze}{HTML}{F0D9C1}

\definecolor{rank_red}{HTML}{FFB3B3}
\definecolor{rank_orange}{HTML}{FFD9B3}
\definecolor{rank_yellow}{HTML}{FFFFB3}

\definecolor{green1}{HTML}{55a630}
\definecolor{green2}{HTML}{c1fba4}
\definecolor{green3}{HTML}{FFFFB3}

\usepackage{bbm}

%% file: sections/abstract.tex
\begin{abstract}

    Reconstructing posed 3D human models from monocular images has important applications in the sports industry, including performance tracking, injury prevention and virtual training.
    In this work, we combine 3D human pose and shape estimation with 3D Gaussian Splatting (3DGS), a representation of the scene composed of a mixture of Gaussians.
    This allows training or fine-tuning a human model predictor on multi-view images alone, without 3D ground truth.
    Predicting such mixtures for a human from a single input image is challenging due to self-occlusions and dependence on articulations, while also needing to retain enough flexibility to accommodate a variety of clothes and poses.
    Our key observation is that the vertices of standardized human meshes (such as SMPL) can provide an adequate spatial density and approximate initial position for the Gaussians. We can then train a transformer model to jointly predict comparatively small adjustments to these positions, as well as the other 3DGS attributes and the SMPL parameters.
    We show empirically that this combination (using only multi-view supervision) can achieve near real-time inference of 3D human models from a single image without expensive diffusion models or 3D points supervision, thus making it ideal for the sport industry at any level.
    More importantly, rendering is an effective auxiliary objective to refine 3D pose estimation by accounting for clothes and other geometric variations. The code is available at \href{https://github.com/prosperolo/GST}{https://github.com/prosperolo/GST}. 
\end{abstract}


%% file: sections/introduction.tex
\section{Introduction} \label{sec:intro}
\vspace{-4pt}

\begin{table*}[t]
\centering
\caption{\textbf{Single Image 3D Human Reconstruction Methods}. Comparison of various 3D representation models, highlighting key attributes such as speed, method of obtaining the model, type of 3D representation, usage of diffusion prior, and supervision technique.}\label{tbl:comparison}
\resizebox{0.88\textwidth}{!}{
\tabcolsep=0.15cm
\begin{tabular}{l rlllll}
\toprule
\textbf{Method} & \textbf{Speed} & \textbf{Obtained by} & \textbf{3D Representation} & \textbf{Diffusion Prior} & \textbf{Supervision} \\
\midrule
\textbf{PIFU} \cite{saito2019pifu} & 10 seconds & Inference & SDF & \xmark & Direct 3D \\
\textbf{HumanLRM} \cite{humanlrm2023} & 7 seconds & Inference & NeRF (Triplane) & \cmark & Direct 3D + MV \\
\textbf{SiTH} \cite{ho2024sith} & 2 {\color{red}minutes} & Inference & SDF & \cmark & Direct 3D \\
\textbf{SIFU} \cite{SIFU} & 6 {\color{red}minutes} & Optimization & SDF & \cmark & Direct 3D \\
\textbf{GTA} \cite{zhang2023globalcorrelated} & 0.55 seconds & Optimization & SDF & \xmark & Direct 3D \\
\textbf{SHERT} \cite{zhan2024shert} & 23 seconds & Inference & Mesh & \cmark & Direct 3D  \\
\textbf{R2Human} \cite{feng2023r2human} & 0.04 seconds & Inference & NeRF (MLP) & \xmark & Direct 3D \\
\textbf{ConTex-Human} \cite{gao2023contexhuman} & 60 {\color{red}minutes} & Optimization & NeRF+Mesh  & \cmark & Direct 3D \\ 
\textbf{SHERF} \cite{SHERF} & 0.75 seconds & Inference & NeRF (MLP) & \xmark & Multi-View \\
\midrule
\textbf{\methodname ~(ours)} & \textbf{0.02s seconds} & Inference & Gaussian Splatting & \xmark & Multi-View \\
\bottomrule
\end{tabular}
}
\end{table*}

Creating posed 3D human models from monocular images is crucial for the sports industry since reliably detecting the shape and pose of all humans in the scene is a fundamental prerequisite for any system monitoring players' health or performance. A reliable system monitoring players' poses throughout different games could be used to analyze their skills training, monitor return to play after injury, and influence game regulations to make the sport safer. These products require precise 3D rendering, speed, compactness, and flexibility to be practical and economically viable for teams at any level. 


Popular approaches like HMR2 \cite{4dhumans} regress the parameters of a human body model such as SMPL \cite{SMPL}. However, these approaches require ground truth 3D pose and shape annotations for every training frame. Therefore, they require significant time and economic investment for data collection before training in any new domain. This makes these methods unsuitable for training in any specific sports setup. Moreover, SMPL can only model the shape of the human body and cannot account for clothing or hair deformations. 


Modeling intricate 3D details and deformations of facial features, clothing, and hair is a long-standing challenge for deep learning methods. Early methods addressed these challenges using a learned Signed Distance Function (SDF) on a human template to predict detailed 3D meshes \cite{saito2019pifu,zhang2023globalcorrelated}. Later works incorporated Neural Radiance Fields (NeRFs) to capture texture details \cite{SHERF,humanlrm2023} or leveraged pre-trained diffusion models to generate dense views from a single frontal image, reducing prediction ambiguity \cite{humanlrm2023, ho2024sith, SIFU, zhan2024shert, gao2023contexhuman, chen2024ultraman}. Despite the good visual quality of the results, these methods suffer from low speed, hindering real-time deployment, and often require expensive 3D scans as supervision.

In this work, we present \methodname (\longmethodname), illustrated in \cref{fig:enter-label}, a direct method that learns to predict 3D Gaussian Splatting \cite{gaussiansplatter} as 3D representation, allowing for fast rendering and flexible editing abilities compared to others. Our method does not rely on diffusion priors and is, therefore, capable of near real-time predictions. 
\methodname leverages multi-view supervision instead of precise (and expensive) 3D point clouds. Despite this, it predicts accurate 3D joint and body poses while maintaining the perceptual quality of renderings from novel views.
\Cref{tbl:comparison} summarizes the characteristics of prior work.

\methodname is inspired by recent works on single-view 3D reconstruction \cite{splatter_image}. However, the complexity of human pose in 3D space poses significant challenges to the direct applications of methods that associate one 3D point (or Gaussian) to each pixel. Therefore, we also augment our model to predict the pose parameters of the SMPL \cite{SMPL} model. The SMPL model is used as the \emph{scaffolding} on which the Gaussians are positioned and rendered. Each Gaussian is loosely tied to a vertex on the SMPL model by an offset. This has two advantages. First, it provides a good initialization of the density and pose of the Gaussians, including back faces, which are notoriously difficult for single-view methods. Second, we find that the joint optimization of pose and appearance improves the SMPL pose prediction. 

To the best of our knowledge, \methodname is the first work that efficiently combines fast and accurate 3D pose human prediction with improved geometry, utilizing only multi-view supervision and without relying on diffusion priors.
In summary, our contributions are the following: 

\noindent\textbf{Contributions:} 
\textbf{(i)} We propose \methodname, a 3D human body model prediction method that does not rely on diffusion priors and performs novel view synthesis and human pose estimation from a single image input. This makes it particularly amenable to real-time modeling tasks, where multiple views are uneconomical or impractical. 
\textbf{(ii)} We evaluate our method and compare it to other state-of-the-art models. In contrast to prior methods that only solve for 3D pose estimation or 3D reconstruction, our method can predict both \textit{without 3D supervision}, making it suitable for training and deployment on any specific sports setup.

%% file: sections/relatedwork.tex
    \begin{figure*}[t]
    \centering
    \includegraphics[trim={0 0 0.2cm 4.0cm},clip,width=0.85\textwidth]{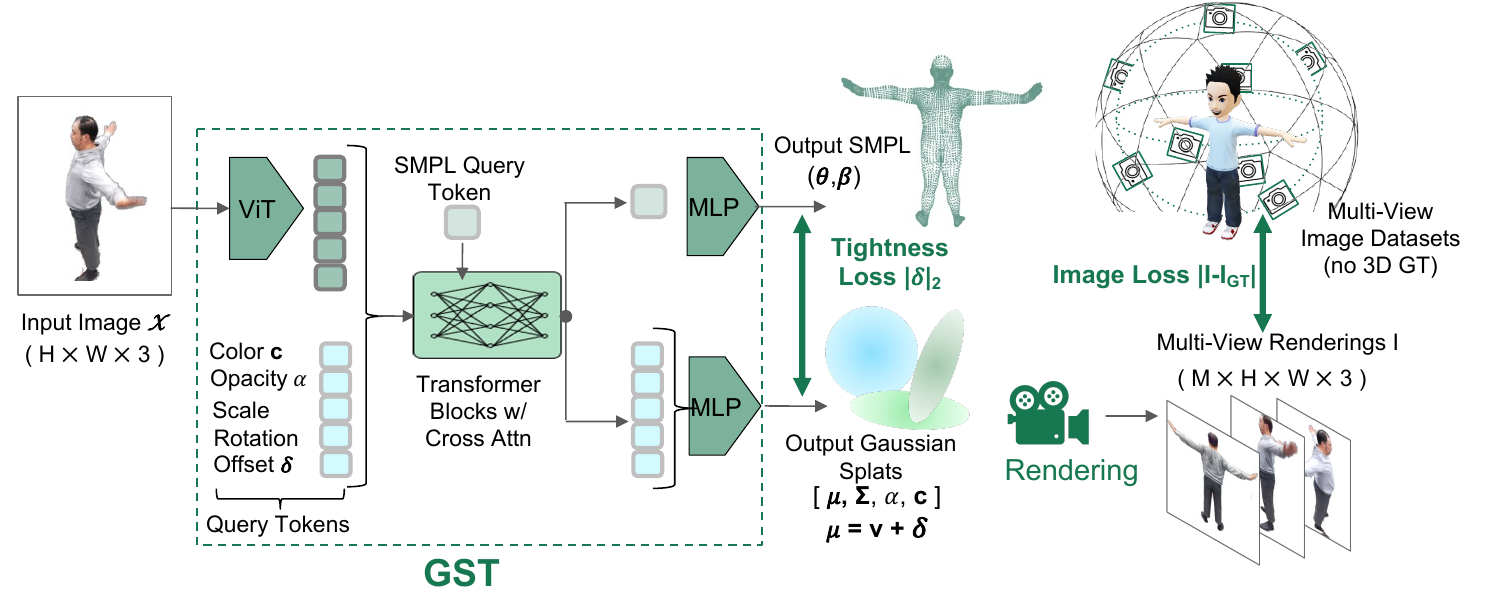}
    \caption{\small \textbf{Overview of the pipeline of \longmethodname{} (\methodname)}. Given a single input image, \methodname uses a Vision Transformer (ViT) to predict both the 3D human pose (in the form of SMPL parameters) and a refined full-color 3D model (in the form of 3D Gaussian Splats). Additional input tokens are used to predict each Gaussian color $\mathbf{c}$, opacity $\alpha$, scale, rotation, and position offset $\boldsymbol{\delta}$. Every Gaussian position $\boldsymbol{\mu}$ is tied to one vertex of the SMPL model $\mathbf{v}$ by the offset $\boldsymbol{\delta}$.
}
    \label{fig:enter-label}
\end{figure*}

\section{Related work}\label{sec:related}
\vspace{-4pt}

\minorsection{3D Human Pose Estimation}
Many approaches in the literature focus on predicting 3D human pose and shape from a single image \cite{kanazawa2018end,kolotouros2019convolutional,lin2021end,cho2022cross,lin2021mesh,4dhumans}. The approaches that directly regress the body shape and pose from a single image are most relevant to our work. The first work to introduce this approach was HMR \cite{kanazawa2018end}, which uses a CNN to regress SMPL \cite{SMPL} parameters. Dedicated designs have been proposed for the HMR architecture; HoloPose~\cite{guler2019holopose} suggests a pooling strategy based on the 2D locations of body joints, while HKMR~\cite{georgakis2020hierarchical} relies on SMPL hierarchical structure to make predictions.  PARE~\cite{kocabas2021pare} introduces a body-part-guided attention mechanism to handle occlusions better, and PyMAF~\cite{zhang2021pymaf,zhang2022pymaf} incorporates a mesh alignment module for SMPL parameter regression. More recently, HMR2 \cite{4dhumans} utilizes a transformer to predict the SMPL parameters and train on a large pool of 3D data and unprotected 2D joint labels. In contrast, TokenHMR \cite{tokenhmr} improved HMR2 by leveraging tokenized encoding and reduced the 2D basis in training HMR2. In contrast to all these methods, our \methodname does not rely on 3D supervision and utilizes novel view synthesis training of a transformer to predict the Gaussian splats grounded on predicted SMPL parameters. Similar to our method, A-NeRF \cite{anerf} jointly optimizes human pose and 3D reconstruction. However, it takes videos as input and does not generalise to unseen subjects.
 

\minorsection{Single-image Human Reconstruction}
Recent advancements in 3D human reconstruction from single images have resulted in diverse methods, each employing different data sources, 3D representations, and supervision strategies \cite{humanlrm2023,SHERF,ho2024sith,SIFU,zhang2023globalcorrelated,zhan2024shert,feng2023r2human,gao2023contexhuman,chen2024ultraman,pesavento2024anim,gaussavatar,gauhuman}. PIFU \cite{saito2019pifu} is one of the earlier works that successfully uses the learned Sign Distance Function (SDF) representation with direct 3D supervision to reconstruct a detailed 3D mesh of humans from a single image. ANIM \cite{pesavento2024anim} incorporates sparse voxel depth features with the input image features and uses direct 3D supervision to train on the RGBD input (depth is needed). SHERF \cite{SHERF} developed on PIFU's 3D representation and adopted a NeRF representation for decoding the 3D human, training with multi-view supervision. While \methodname follows SHERF in the multi-view supervision, we utilize the more explicit Gaussian Splatting representation \cite{gaussiansplatter} for 3D, allowing for more flexible control and better 3D alignment. 

With the recent wave of success of generative image and text models ~\cite{DALL-E, LDM, DALLE-2, Imagen, GPT4}, several methods try to leverage these foundation models to improve the performance of 3D human reconstruction~\cite{SIFU,ho2024sith,humanlrm2023,feng2023r2human,chen2024ultraman,gao2023contexhuman}. For example, SIFU~\cite{SIFU} integrates GPT-predicted captions with diffusion models for back-view generation and texture refinement and builds on GTA~\cite{zhang2023globalcorrelated}, its predecessor, which learns a triplane SDF decoder. Similarly, SiTH \cite{ho2024sith} employs a diffusion prior to generate back views and decode the SDF and texture colours, while HumanLRM \cite{humanlrm2023} generates multi-views with a pre-trained diffusion model and then trains a tri-plane NeRF Large Reconstruction Model (LRM) for decoding. ELICIT~\cite{elicit} leverages CLIP~\cite{CLIP} to generate text-conditioned unseen regions. SHERT \cite{zhan2024shert} utilizes semantic mesh and whole texture inpainting with the help of diffusion priors to create detailed 3D meshes of humans. HumanSplat~\cite{pan2024humansplat} and Human 3diffusion~\cite{human3diffusion} use diffusion priors to generate novel views, and TeCH~\cite{tech} uses text-to-image diffusion models to embed indescribable appearance information. Most of these methods rely on pre-trained diffusion models for texturing optimization, which slows down the process and limits scalability for high-speed, long-duration video motion. In contrast, our \methodname operates without diffusion priors, achieving near real-time inference while remaining flexible and capable of integrating with priors if needed.

%% file: sections/methodology.tex
\section{Gaussian Splatting Transformers (GST)}\label{sec:method}
\vspace{-4pt}
This section presents our methodology for reconstructing 3D human models from a single image using Gaussian Splatting Transformers (GST), as illustrated in \cref{fig:enter-label}.

\subsection{Architecture}\label{sec:arch}
Our model predicts 3D Gaussian splatting parameters from a single input image using a transformer architecture, including tokenization, processing through transformer blocks, and decoding into Gaussian parameters.
We detail the model architecture (\cref{sec:arch}) and the loss functions (\cref{sec:loss}).

\minorsection{Image Encoder Architecture}
Our backbone follows HMR2 \cite{4dhumans} and uses a ViT \cite{ViT} to map an image to a series of visual tokens.
The input is an RGB image $\mathbf{X} \in \mathbb{R}^{H \times W \times 3}$, which is divided into non-overlapping patches $\mathbf{p}_j \in \mathbb{R}^{p \times p \times 3}$, with  $j\in\{1,\ldots,HW/p^{2}\}$. The patches are vectorized and affinely transformed into patch tokens $\mathbf{x}_j \in \mathbb{R}^{d}$.

The patch tokens are processed through a series of Transformer blocks \cite{vaswani2017attention}. The final output is a set of tokens $\mathbf{y}_j \in \mathbb{R}^{d}$ encapsulating the transformed image information.

\minorsection{Human Shape Representation}
The SMPL model \cite{SMPL} represents the 3D human mesh shape as a mesh. 
SMPL is a low-dimensional parametric model defined by pose parameters $\boldsymbol{\theta} \in \mathbb{R}^{24 \times 3 \times 3}$ and shape parameters $\boldsymbol{\beta} \in \mathbb{R}^{10}$, outputting mesh vertices' 3D positions $\mathbf{v} = \text{SMPL}(\boldsymbol{\theta}, \boldsymbol{\beta}) \in \mathbb{R}^{6890 \times 3}$.

\minorsection{Decoder Architecture}
We build on HMR2 \cite{4dhumans}, which predicts the SMPL representation $(\boldsymbol{\theta}, \boldsymbol{\beta})$ from the image representation $\mathbf{y}_j$ through a cross-attention mechanism. Specifically, a single (fixed) token $t_\mathrm{SMPL}$ attends to all image tokens $\mathbf{y}_j$ through a series of cross-attention layers. 
An MLP decodes the token into the pose parameters $\boldsymbol{\theta}$ and $\boldsymbol{\beta}$.

This representation could be learned with image-pose pairs $(\mathbf{X}, \boldsymbol{\theta}, \boldsymbol{\beta})$. However, here, we focus on multi-view supervision, as 3D supervision is costly and scarce, and not readily available for real-world setups. 

To train with multi-view supervision, the model needs to generate an image. 
We use recent advances in fast neural rendering: Gaussian Splatting \cite{gaussiansplatter}.
This scene representation is defined by a set of 3D Gaussians, each characterized by a mean position $\boldsymbol{\mu} \in \mathbb{R}^3$, a covariance matrix $\boldsymbol{\Sigma} \in \mathbb{R}^{3 \times 3}$, the opacity $\alpha \in \mathbb{R}$ and a colour $\mathbf{c} \in \mathbb{R}^3$. 

We link the 3D body shape and pose with the Gaussian scene representation, such that each vertex $\mathbf{v}_n$ in the mesh is assigned a Gaussian $G_n = (\boldsymbol{\mu}_n, \boldsymbol{\Sigma}_n, \alpha_n, \mathbf{c}_n)$.
We allow the Gaussians to move away from the original vertex positions by a learned offset $\boldsymbol{\delta}_n$ to model clothes and other visual shape features that the SMPL model cannot capture.
\begin{align}
    \boldsymbol{\mu}_n = \mathbf{v}_n + \boldsymbol{\delta}_n, \label{eq:smpl_gaussian}
\end{align}
This combination ensures that the 3D model captures both shape and appearance, allowing for more realistic reconstructions.

Similar to prior work~\cite{splatter_image}, we factorize and simplify the covariance into the product of a rotation matrix and a diagonal matrix, enforcing a reduced number of degrees of freedom from 9 to 6: $G_n \in \mathbb{R}^{14}$. 

It is theoretically possible to assign five tokens per Gaussian, one for each parameter: rotation, offset, scale, color, and opacity. However, this would result in over 34k tokens, which is computationally infeasible to decode with a standard Transformer.
We thus group vertices into $K$ groups, reducing the number of tokens to $5K + 1$ (in practice, we set $K=26$). As discussed before, the additional token is used to predict the SMPL shape parameters.  

This representation allows initialization with the pre-trained weights of HMR2 \cite{4dhumans} since we only introduce additional (fixed but learned) tokens in the decoder architecture.
A set of Gaussians can be assembled from the predictions, which can be rendered into an image from any viewpoint. 


\subsection{Loss Functions}\label{sec:loss}
\vspace{-2pt}
We use a combination of losses to train our model to ensure accurate and visually realistic 3D reconstructions.

\minorsection{Image Reconstruction Loss}
We use a combination of  Mean Squared Error (MSE) to measure the difference between the $M$ multi-view ground truth images $\hat{\mathbf{I}}_i$ and rendered images $\mathbf{I}_i$, a perceptual loss to capture high-level features and textures with LPIPS metric \cite{LPIPS}, and a masked loss on the rendered opacity $\mathbf{I}_{i}^{\alpha}$ to remove background splats \cite{gslrm}:
\begin{equation}     
\begin{aligned}
\mathcal{L}_{\text{img}} = \frac{1}{M} \sum_{i=1}^M \Big( & \left\| \hat{\mathbf{I}}_i - \mathbf{I}_i \right\|_2^2 +  \lambda_{\text{perceptual}} \cdot \text{LPIPS}(\hat{\mathbf{I}}_i, \mathbf{I}_i) \\
    & + ~ \lambda_{\alpha} \left\| \hat{\mathbf{M}}_{i} - \mathbf{I}_{i}^{\alpha} \right\|_2^2 ~\Big),
\end{aligned}
\label{eq:img_loss}
\end{equation}
where $\hat{\mathbf{M}}_{i}$ is the background mask of the ground truth images $\hat{\mathbf{I}}_i$, and $\lambda_{\text{perceptual}}$ and  $\lambda_{\alpha}$ are weighting hyperparameters for the perceptual and transparency losses respectively. The transparency loss is necessary to reduce floating Gaussians that do not contribute to the foreground object.  

\minorsection{Gaussian Tightness Regularization}
To ensure that the predicted Gaussian Splats in  \cref{sec:arch} follow the SMPL parameters closely, we introduce a Gaussian tightness regularization that ensures the generated Gaussian splats \cite{gaussiansplatter} do not diverge and remain faithful to the underlying SMPL parameters as follows:
\begin{align}
    \mathcal{L}_{\text{tight}} = \frac{1}{V} \sum_{n=1}^{V} \left\| \boldsymbol{\delta}_n \right\|_2,
    \label{eq:tight_loss}
\end{align}
where $\boldsymbol{\delta}_n$ is defined in \eqref{eq:smpl_gaussian} and $V=6890$ is the number of Gaussian splats (number of vertices in SMPL).


The total loss function is a weighted sum of the image losses (MSE, perceptual, and alpha) and tightness:
\begin{align}
    \mathcal{L} = \mathcal{L}_{\text{img}} + \lambda_{\text{tight}} \mathcal{L}_{\text{tight}},
    \label{eq:total_loss}
\end{align}
where $\lambda_{\text{tight}}$ is the weighting hyperparameter for the tightness regularization.
As we show later in \cref{sec:ablation}, this regularization plays an important role in the precision of the 3D human body predicted by \methodname. By minimizing this combined loss, our \methodname model learns to generate accurate human 3D models from a single image. 

Optionally, a further regularization term for the SMPL shape parameters can also be included. This regularization term is an L2 loss on the $\beta$ parameters. This additional term does not affect the pose or visual metrics. It only influences the predicted body thickness.

%% file: sections/experiments.tex
\section{Experiments}\label{sec:exp}
\vspace{-4pt}
This section describes our evaluation setup and the baselines and prior work used in our comparisons.

\subsection{Datasets and Metrics}\label{sec:dataset}
\minorsection{Datasets}
Similar to previous works \cite{SHERF}, we utilize four comprehensive human datasets for evaluation: THuman~\cite{Thuman}, RenderPeople~\cite{renderpeople}, ZJU MoCap~\cite{ZJUMoCap}, and HuMMan~\cite{humman}. For ZJU MoCap, the dataset is divided following the SHERF setup~\cite{SHERF}. Similarly, for HuMMan, we adhere to the official split (HuMMan-Recon), using 317 sequences for training and 22 for testing, with 17 frames sampled per sequence. For THuman, we select 90 subjects for training and 10 for testing, and for the RenderPeople dataset, we randomly sample 450 subjects for training and 30 for testing. Those four datasets used for evaluation above are all small in terms of subject diversity. To showcase the capabilities of \methodname on a large dataset, we train our \methodname also on the TH21 dataset \cite{th2}, which contains 2,500 3D scans with high subject diversity. We randomly selected 200 scans for evaluation. Further validation is also conducted on the CMU Panoptic dataset with the single human partition that includes 9 sequences with 31 HD camera views; some qualitative examples of the pose estimation results for the sports sequence in this dataset are shown in \cref{fig:teaser} and in \supp. Another dataset that could have been suitable for evaluating our method is the SportsPose dataset \cite{sportspose}; however, no multi-view data has been released. 

\begin{table*}[t]
\centering
\caption{\textbf{Human Novel View Synthesis and 3D Keypoints Evaluation Performance Comparison.} We compare \methodname on the RenderPeople \cite{renderpeople} and HuMMan \cite{humman} datasets. We report PSNR, SSIM, and LPIPS for novel view synthesis and MPJPE (in mm) for 3D keypoint evaluation for each dataset. The top section methods use the Ground Truth input SMPL parameters, shown for reference, while the bottom section methods only use the single image input (our setup).  
$\uparrow$ means the larger is better; $\downarrow$ means the smaller is better. The best results are highlighted in \textbf{bold}.
}
\resizebox{0.93\linewidth}{!}{
\small
\setlength{\tabcolsep}{1.3mm}{
\begin{tabular}{lcc cccc cccc}
\toprule
\multirow{3}*{Method} & &  & \multicolumn{4}{c}{\textit{RenderPeople}} & \multicolumn{4}{c}{\textit{HuMMan}} \\
\cline{4-7} \cline{8-11}
& \multicolumn{2}{c}{GT 3D} & \multicolumn{3}{c}{Novel View} & 3D Shape & \multicolumn{3}{c}{Novel View} & 3D Shape \\
& Test & Train & PSNR$\uparrow$ & SSIM$\uparrow$ & LPIPS$\downarrow$ & MPJPE (mm)$\downarrow$ & PSNR$\uparrow$ & SSIM$\uparrow$ & LPIPS$\downarrow$ & MPJPE (mm)$\downarrow$ \\
\hline
NHP~\cite{kwon2021neural} & \cmark & \cmark & 20.59 & 0.81 & 0.22  & 0.000 & 18.99 & 0.84 & 0.18 & 0.000 \\
MPS-NeRF~\cite{gao2022mps} & \cmark & \cmark & 20.72 & 0.81 & 0.24 & 0.000 & 17.44 & 0.82 & 0.19 & 0.000 \\
SHERF~\cite{SHERF} w/ GT & \cmark & \cmark & 22.88 & 0.88 & 0.14  & 0.000 & {20.83} & {0.89} & {0.12} & 0.000 \\
HMR2~ (3D fine-tuned)& \xmark & \cmark & - & - & - & {57.33} & - & - & - & {61.20} \\
\midrule\midrule
HMR2~\cite{4dhumans}& \xmark & \xmark & - & - & - & 101.0 & - & - & - & 133.4 \\
HMR2~ (2D-only fine-tuned)& \xmark & \xmark & - & - & - & 127.40 & - & - & - & 163.77 \\
TokenHMR \cite{tokenhmr} & \xmark & \xmark & - & - & - & 77.9 & - & - & - & 91.4 \\
SHERF~\cite{SHERF} w/~\cite{4dhumans} & \xmark &  \xmark & 13.55 & 0.62 & 0.37  & 101.0 & {18.00} & {0.85} & 0.18 & 133.4 \\
SHERF~\cite{SHERF} w/~\cite{tokenhmr} & \xmark & \xmark & {15.24} & {0.70} & {0.33}  & 77.9 & 16.41 & 0.84 & {0.17} & 91.4 \\
\textbf{\methodname (Ours)} & \xmark & \xmark & \textbf{17.80} & \textbf{0.81} & \textbf{0.25} & \textbf{67.6} & \textbf{18.40} & \textbf{0.87} & \textbf{0.14} & \textbf{64.6} \\
\bottomrule
\end{tabular}}
}
\label{tab:3d_result}
\end{table*}

\minorsection{Evaluation Metrics}
When the Ground Truth 3D SMPL parameters are available as in RenderPeople \cite{renderpeople} and HuMMan \cite{humman}, we adopt 3D Human Joints precision MPJPE as a metric \cite{kanazawa2018end}.
MPJPE refers to Mean Per Joint Position Error: the average L2 error across all joints after aligning with the root node. 
To quantitatively assess the quality of rendered novel view and novel pose images, we report peak signal-to-noise ratio (PSNR), structural similarity index (SSIM), and Learned Perceptual Image Patch Similarity (LPIPS)\cite{LPIPS}. Consistently with prior works\cite{Thuman, gao2022mps,SHERF}, we project the 3D human bounding box onto each camera plane to derive the bounding box mask, subsequently reporting these metrics based on the masked regions. 

\minorsection{Baselines}
In addition to earlier work on Human NeRF with a multi-view setting, NHP~\cite{kwon2021neural} and MPS-NeRF~\cite{gao2022mps}, we compare to recent single-image methods SHERF \cite{SHERF} for novel view synthesis and HMR2 \cite{4dhumans} and TokenHMR \cite{tokenhmr} for 3D Human reconstruction precision. Unlike SHERF, our method does not take as input ground truth SMPL parameters. Therefore, we adapt SHERF to use HMR2 \cite{4dhumans} or TokenHMR \cite{tokenhmr} SMPL predictions for a fair comparison to our method. We also include a comparison with Splatter Image \cite{splatter_image}, a state-of-the-art single-image 3D reconstruction method for novel-view synthesis tables.

\begin{figure*}[h]
    \begin{tikzpicture}[font=\footnotesize]
\node (A) [anchor=west] at (0,0) {\includegraphics[width=0.95\textwidth]{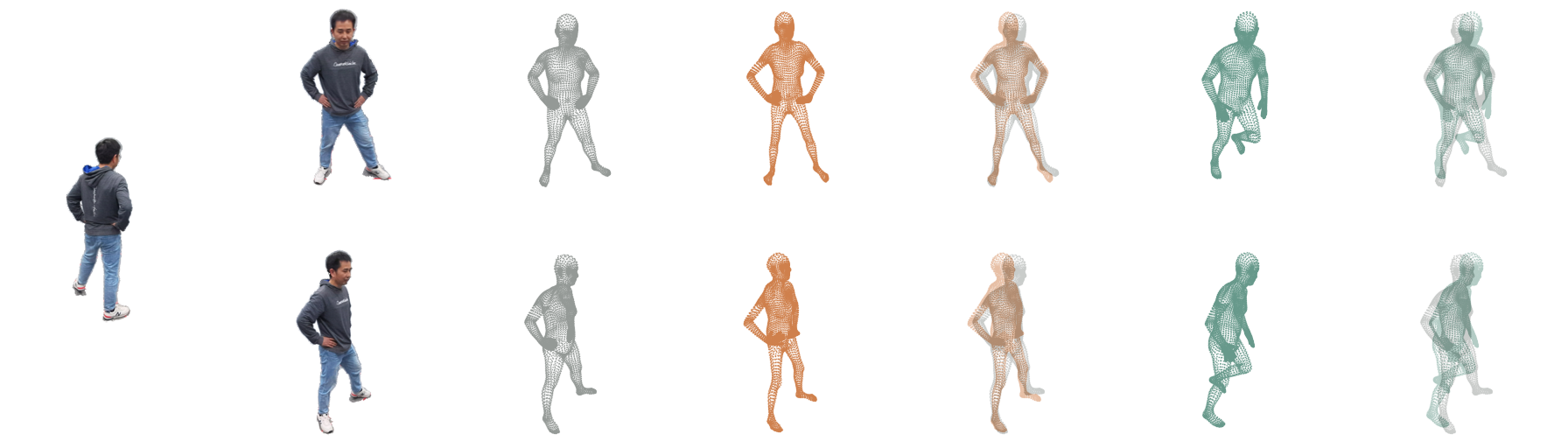}};
\node (D) [anchor=north] at (A.south) {\includegraphics[width=0.95\textwidth]{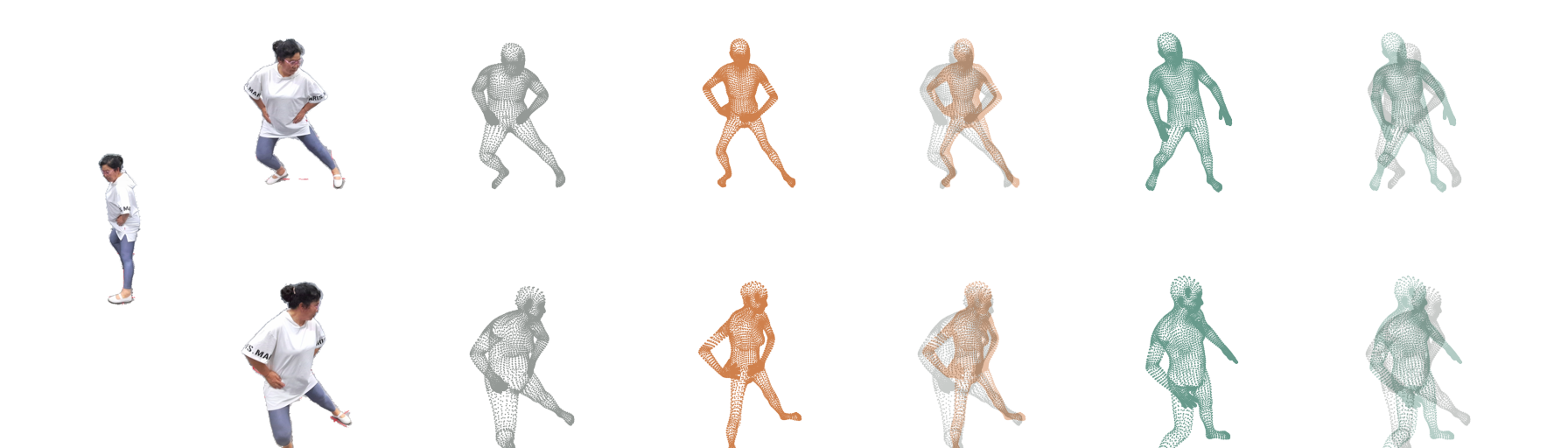}};
\node (B) [anchor=west] at ([xshift=-8cm,yshift=0.5cm] A.north) {Input Image};
\node (C) [anchor=west] at ([xshift=-5.5cm,yshift=0.5cm] A.north) {Other Views};
\node (D) [anchor=west] at ([xshift=-2.8cm,yshift=0.5cm] A.north) {GT};
\node (E) [anchor=west] at ([xshift=-0.4cm,yshift=0.5cm] A.north) {\textbf{GST (ours)}};
\node (H) [anchor=west] at ([xshift=0.8cm,yshift=-0.1cm] E.south) {vs GT};
\node (F) [anchor=west] at ([xshift=4.2cm,yshift=0.5cm] A.north) {HMR2 \cite{4dhumans}};
\node (G) [anchor=west] at ([xshift=0.8cm,yshift=-0.1cm] F.south) {vs GT};
    \end{tikzpicture}
    \hfill
    \caption{\textbf{3D Shape Comparison with HMR2.} 3D human body results of our \methodname on two subjects of HuMMan \cite{humman} dataset compared to Ground Truth SMPL parameters \cite{SMPL}, and SMPL predictions of HMR2 \cite{4dhumans}.
}
    \label{fig:3dresults}
\end{figure*}

\subsection{Implementation Details of \methodname} \label{sec:details}
Our model follows the implementation of HMR2 \cite{4dhumans} for the prediction of the SMPL parameters. We extend the HMR2 decoder implementation to process some additional learnable tokens for the predictions of the Gaussian parameters. The Gaussian parameters (color, rotation, scale, opacity, and offset) are predicted for $K=26$ groups of 265 Gaussians. The token output is passed through a linear layer to obtain the final parameters. We use pre-trained HMR2 weights for the ViT and the decoder and freeze the ViT weights during training. For our experiments, we use the loss weights $\mathcal{L}_{\text{perceptual}} = 0.01$, $\mathcal{L}_\alpha = 0.1$, $\mathcal{L}_{\text{tight}} = 0.1$, and we train on square image crops of size 256. We train on a single A6000 GPU with a batch size of 32 for 3 days. At test time, \methodname can simultaneously perform 3D human pose estimation and 3D reconstruction in a single forward pass at 47fps. We verify that the network can learn to predict high-quality renderings by overfitting on a single dataset example (\cf \supp for more details). 

\begin{table}[t]
\addtolength{\tabcolsep}{5pt}
\centering
\caption{\textbf{Novel View Synthesis Performance Comparison.} We compare \methodname on the ZJU\_MoCap \cite{ZJUMoCap} and THuman \cite{Thuman} datasets on novel view synthesis. The top section methods use the Ground Truth input SMPL parameters (allowing for changing the pose) and are shown for reference, while the bottom section methods only use the single image input (our setup).  }
\label{tab:2d_result}
\resizebox{0.99\linewidth}{!}{
\small
\setlength{\tabcolsep}{1.0mm}{
\begin{tabular}{l ccc ccc}
\toprule
\multirow{2}*{Method} & \multicolumn{3}{c}{\textit{ZJU MoCap}} & \multicolumn{3}{c}{\textit{THuman}} \\ \cline{2-7} 
 & PSNR$\uparrow$ & SSIM$\uparrow$ & LPIPS$\downarrow$ & PSNR$\uparrow$ & SSIM$\uparrow$ & LPIPS$\downarrow$ \\
\midrule
PixelNeRF~\cite{PixelNeRF} & - & - & - & 16.51 & 0.65 & 0.35 \\
NHP~\cite{kwon2021neural} &  21.66 & 0.87 & 0.17 & 22.53 & 0.88 & 0.17  \\
MPS-NeRF~\cite{gao2022mps} &  21.86 & 0.87 & 0.17 & 21.72 & 0.87 & 0.18  \\
SHERF~\cite{SHERF} /w GT & 22.87 & 0.89 & 0.12 & 24.66 & 0.91 & 0.10  \\
\midrule\midrule
Splatter Img~\cite{splatter_image} & 19.50 & 0.80 & 0.28  & 19.20 & 0.80 & 0.20 \\
SHERF~\cite{SHERF} /w~\cite{4dhumans} & 19.11 & 0.81 & 0.21 & 17.27 & \textbf{0.85} & \textbf{0.16} \\
SHERF~\cite{SHERF} /w~\cite{tokenhmr} & 20.72 & 0.85 & 0.16  & \textbf{19.29} & 0.84 & 0.18 \\
\textbf{\methodname (Ours)} & \textbf{21.26} & \textbf{0.85} & \textbf{0.16} & 16.34 & 0.84 & 0.20 \\
\bottomrule
\end{tabular}}
}
\end{table}

%% file: sections/results.tex
\section{Results}\label{sec:results}
\vspace{-4pt}
In this section, we discuss the results obtained on five datasets in various evaluation settings.

\subsection{3D Human Shape Results}
The primary focus of this work is the ability to infer a precise 3D human body from a single image without explicit 3D supervision. We show quantitative results in \cref{tab:3d_result} on RenderPeople \cite{renderpeople} and HuMMan datasets \cite{humman}. We compute the MPJPE error with respect to the ground truth SMPL joints before and after training. The results show that without explicit 3D supervision, our training improves the quality of the pose estimation from the pretrained HMR2 \cite{4dhumans} and TokenHMR \cite{tokenhmr}. Furthermore, \cref{fig:3dresults} shows some examples of our predictions in comparison with the HMR2 initialization and the ground truth SMPL pose. Our poses appear visually to be better aligned with the ground truth, highlighting the results in \cref{tab:3d_result}. 

\begin{figure*}[h]
    \begin{tikzpicture}[font=\footnotesize]
    \centering
\node (A) [anchor=west] at (0,0) {\includegraphics[width=0.95 \textwidth]{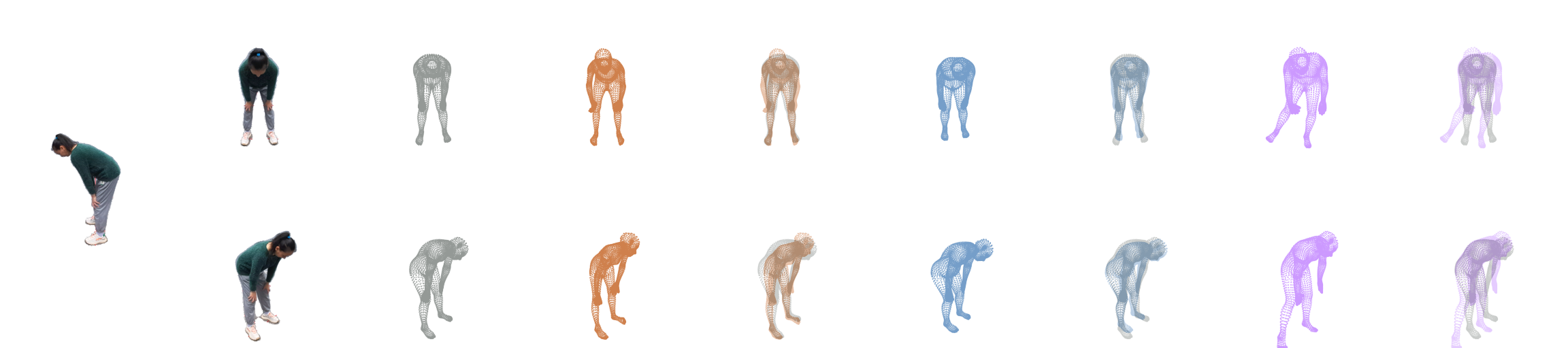}};
\node (D) [anchor=north] at (A.south) {\includegraphics[width=0.95 \textwidth]{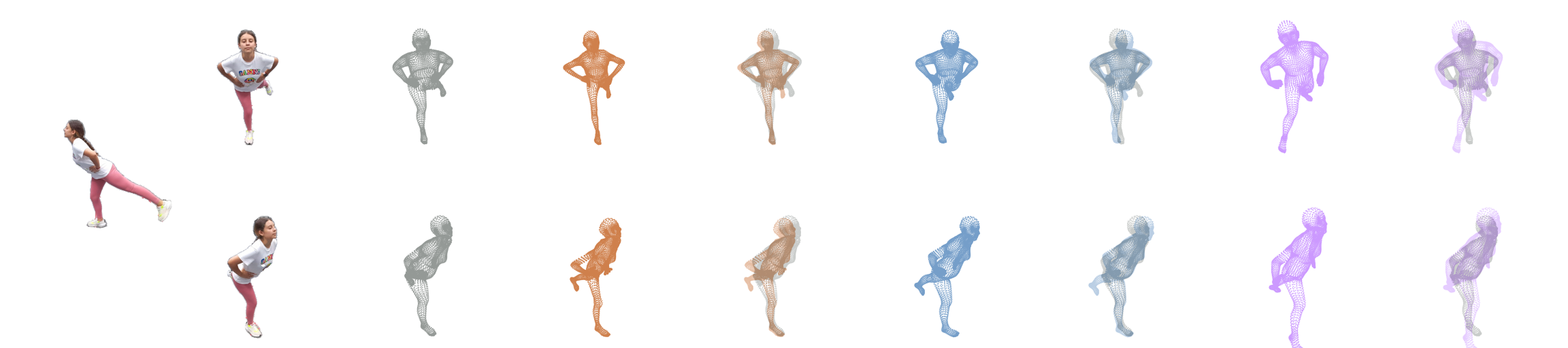}};
\node (B) [anchor=west] at ([xshift=-8.5cm] A.north) {Input Image};
\node (C) [anchor=west] at ([xshift=-6.2cm] A.north) {Other Views};
\node (D) [anchor=west] at ([xshift=-4cm] A.north) {GT};
\node (E) [anchor=west] at ([xshift=-1.8cm] A.north) {\textbf{GST (ours)}};
\node (H) [anchor=west] at ([xshift=0.5cm, yshift=-0.1cm] E.south) {vs GT};
\node (F) [anchor=west] at ([xshift=1.4cm] A.north) {HMR2-3D finetuning};
\node (I) [anchor=west] at ([xshift=0.5cm, yshift=-0.1cm] F.south) {vs GT};
\node (G) [anchor=west] at ([xshift=5.1cm] A.north) {HMR2-2D finetuning};
\node (J) [anchor=west] at ([xshift=0.5cm, yshift=-0.1cm] G.south) {vs GT};
    \end{tikzpicture}
    \caption{\textbf{3D Shape Comparison with HMR2 After Fine-tuning on 2D and 3D Data.} 3D human body results of our \methodname on two subjects of HuMMan \cite{humman} dataset compared to Ground Truth SMPL parameters \cite{SMPL}, and SMPL predictions of HMR2 \cite{4dhumans}. We show two versions of HMR2, one finetuned on 2D data only (HMR2-2D), and one finetuned on 3D data (HMR2-3D). Our method is only finetuned on 2D image data, but the results are almost as accurate as HMR2 finetuned on 3D data. 
}
    \label{fig:3dresults_finetuned}
\end{figure*}

We also compare our results against a fine-tuned version of HMR2 (c.f.~\cref{tab:3d_result}). To reproduce a similar training setup to our method (that does not require any 3D ground truth annotations), we fine-tune HMR2 using only 2D keypoint annotations. We use images from all views in the dataset but restrict the supervision only to use the 2D keypoints loss. The results show that the 2D information alone is not enough for HMR2 to improve the quality of the 3D pose estimation on the two datasets, and the fine-tuned model MPJPE error is worse than the pretrained one. For completeness, we also report the errors when fine-tuning HMR2 with additional 3D annotations: 3D keypoints and ground truth SMPL parameters. We would like to emphasize that we think this is an unfair comparison to our method since our method does not use ground truth SMPL parameters or 3D keypoints for training. The MPJPE of the HMR2 version, fine-tuned with 3D data, is only 7mm better than ours on RenderPeople and 6mm better than ours on HuMMan. \Cref{fig:3dresults_finetuned} shows some examples of our predictions compared to the HMR2 models finetuned on 2D and 3D data. 


\subsection{Novel View Synthesis Results}
\begin{figure*}[h]
    \begin{tikzpicture}[font=\footnotesize]

\node (A) at (0,0) {\includegraphics[width=0.97\textwidth]{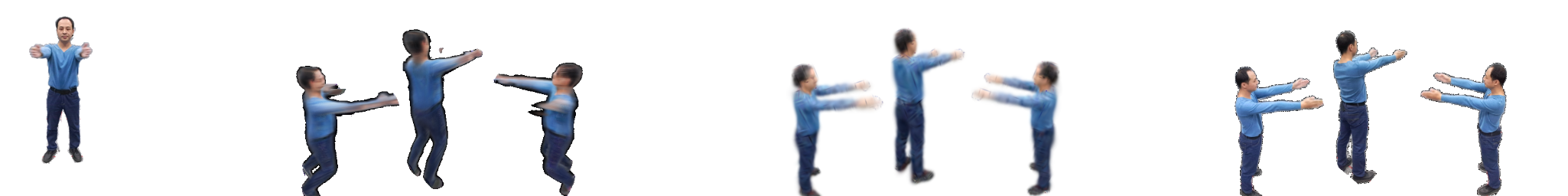}};
\node (D) [anchor=north] at (A.south) {\includegraphics[width=0.97\textwidth]{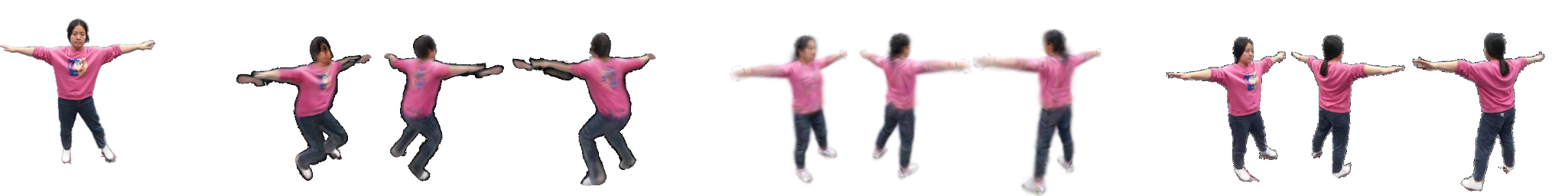}};
\node (B) [anchor=west] at ([xshift=-8.5cm] A.north) {Input Image};
\node (C) [anchor=west] at ([xshift=-5cm] A.north) {SHERF \cite{SHERF} w/ \cite{4dhumans}};
\node (D) [anchor=west] at ([xshift=0.5cm] A.north) {\textbf{GST (ours)}};
\node (D) [anchor=west] at ([xshift=5.8cm] A.north) {GT};
    \end{tikzpicture}
    \caption{\textbf{Single Image NVS}. \methodname on 2 subjects of HuMMan \cite{humman} dataset compared to Ground Truth renderings, and SHERF \cite{SHERF} (after being adapted with HMR2 to work with single image input only). \methodname depicts the \emph{correct} human pose (compared with ground truth).}
    \label{fig:2dresults_humman}
\end{figure*}

We evaluate our method in the task of novel view synthesis across 4 datasets and compare the results with SHERF \cite{SHERF}. For a fair comparison with our method, which does not assume ground truth SMPL parameters are available, we evaluate SHERF using the estimated HMR2/TokenHMR pose and shape parameters instead of the ground truth ones. Our results are in Tables \cref{tab:3d_result,tab:2d_result}. Visual results are shown in Fig. \cref{fig:2dresults_humman}. Note that the underlying 3D body is consistent, despite a slight blurriness due to the Gaussians, and follows precise 3D geometry. 

\begin{figure}
\centering
\includegraphics[width=0.45\textwidth]{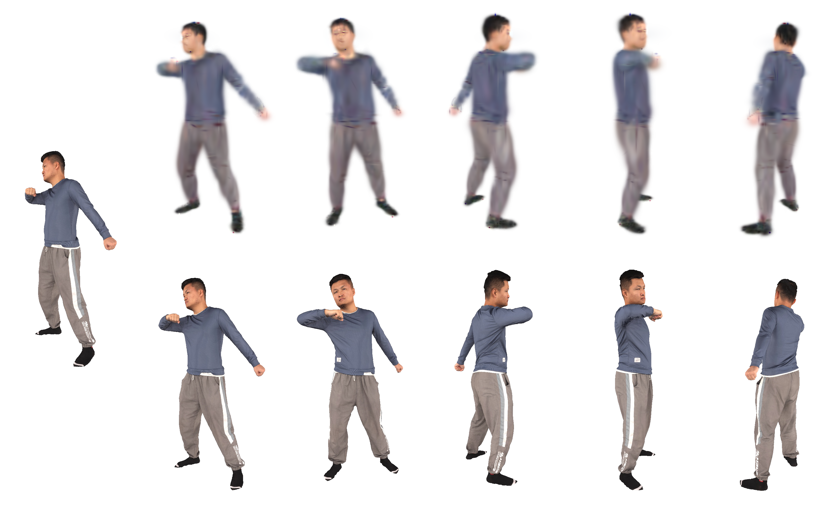}
\caption{\textbf{Scaling Up Training of GST on TH21.} We show rendering results for \methodname (\textit{top row}) compared to Ground Truth renderings (\textit{bottom row}) of each subject. The training of 2,500 subjects in TH21 \cite{th2} reduces the blurriness observed in other datasets and demonstrates the scalability merit of our \methodname Transformer training.} 
\label{fig:thuman2_results}
\end{figure}

To showcase the capabilities of GST on a large dataset, we train our \methodname on multi-view images rendered from the TH21 dataset \cite{th2}, which contains 2,500 3D scans and shows the results on 200 randomly sampled test scans in \cref{tab:thuman2} and \cref{fig:thuman2_results}. It clearly shows less blurriness than the other datasets. For reference, we include Splatter Image \cite{splatter_image} in \cref{tab:thuman2}, where our \methodname predicts precise 3D body pose and shape in addition to the renderable representation, unlike Splatter Image. Predicting the human model parameters is not only useful for downstream tasks but also ensures the reconstructed shape is plausible for a human. We include additional details on the TH21 experiment and the comparison with Splatter Image in \supp{}. 


\begin{table}[t]
\centering
\caption{\textbf{Novel View Synthesis on Large-Scale TH21}. We compare \methodname to fast and large-scale multi-view baseline \cite{splatter_image} that do not need 3D annotations on the 2,500 examples from TH21 \cite{th2}. Unlike Splatter Image \cite{splatter_image}, our \methodname predicts precise 3D body pose and shape in addition to the renderable representation. 
}
\label{tab:thuman2}
\vspace{-0.5em}
\resizebox{0.85\linewidth}{!}{
    \centering
    \small{
    \addtolength{\tabcolsep}{-3.pt}
    \begin{tabular}{l c ccc}
        \toprule
        Method & Output & \multicolumn{3}{c}{Novel View} \\
         ~ & 3D Body & PSNR$\uparrow$ & SSIM$\uparrow$ & LPIPS$\downarrow$  \\
        \midrule
        Splatter Img~\cite{splatter_image} & \xmark &  \textbf{23.74} & \textbf{0.91} & 0.10 \\
        \textbf{\methodname (Ours)} & \checkmark & 22.20 & 0.90 & \textbf{0.09}  \\
        \bottomrule
    \end{tabular}}
}
\end{table}

%% file: sections/ablation.tex
\subsection{Ablation and analysis}\label{sec:ablation}


\begin{figure}
    \centering

\resizebox{1\hsize}{!}{
\begin{tikzpicture}[font=\footnotesize]

\node (A) at (0,0) {\includegraphics[width=0.45\textwidth]{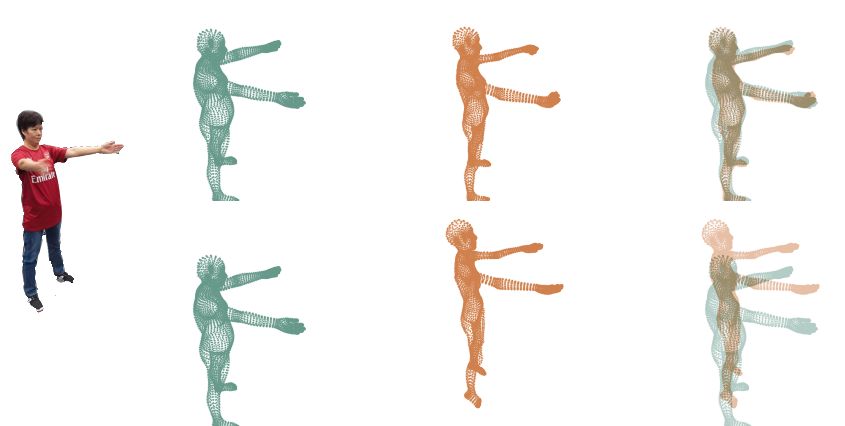}};
\node (B) [anchor=west] at ([xshift=-4.5cm] A.north) {Input Image};
\node (C) [anchor=west] at ([xshift=-2.5cm] A.north) {GT SMPL};
\node (D) [anchor=west] at ([xshift=-0.1cm] A.north) {Our SMPL };
\node (D) [anchor=west] at ([xshift=2cm] A.north) {SMPL overlays};
    \end{tikzpicture}

    }
    \caption{\textbf{Tightness Regularization.} Renderings and SMPL models with (\textit{top}) and without (\textit{bottom}) tightness regularization. The regularization maintains a precise body shape and pose.}
    \label{fig:tightness}
\end{figure}

\begin{table}
\centering
    \caption{
        \textbf{Ablation study}. We show an ablation study on HuMMan Dataset \cite{humman} where the left shows the design choices and the right part shows the results. For each setup, we report PSNR, SSIM, and LPIPS for novel view synthesis, as well as MPJPE (in mm) for 3D keypoints evaluation. 
    } \label{tab:ablation}
\resizebox{0.99\linewidth}{!}{
    \centering
    \small{
    \addtolength{\tabcolsep}{-1.pt}
    \begin{tabular}{ccc cccc}
        \toprule
        \multirow{2}*{\makecell{LPIPS \\ loss}} & \multirow{2}*{\makecell{Tightness \\ loss}} & \multirow{2}*{\makecell{Transparency \\ loss}} & \multicolumn{3}{c}{Novel View} & \multicolumn{1}{c}{3D Shape} \\
        \cmidrule{4-7}
         ~ & ~ & ~ & PSNR$\uparrow$ & SSIM$\uparrow$ & LPIPS$\downarrow$ & MPJPE (mm)$\downarrow$ \\
        \midrule
          $\checkmark$ & & $\checkmark$ & 21.77 & 0.87 & 0.12 & 82.3 \\
          & $\checkmark$ & $\checkmark$ & \textbf{21.80} & 0.86 & 0.15 & 53.6 \\
         $\checkmark$ & $\checkmark$ & & 21.77 & 0.87 & 0.12 & 52.3 \\
        $\checkmark$ & $\checkmark$ & $\checkmark$ & 21.79 & \textbf{0.87} & \textbf{0.12} & \textbf{50.8} \\ 
        \bottomrule
    \end{tabular}}
}
\end{table}
\minorsection{Ablation Study}
We present an ablation study of different design choices and key elements in our architectures and losses and their effect on the 2D and 3D results of a single image to 3D of humans on the HuMMan dataset \cite{humman} in \cref{tab:ablation}. For these experiments, we report PSNR, SSIM, and LPIPS metrics computed for the entire image. It shows the importance of combining the LPIPS, tightness, and transparency loss on the final 3D precision while maintaining the visual fidelity intact.
The tightness regularization of \eqref{eq:tight_loss} has the highest impact on 3D precision, as it favors solutions in which the majority of the pose corrections are obtained with the SMPL parameters, and the Gaussians are only used for small refinements. In contrast, removing the tightness regularization encourages unrealistic and less precise poses, with much larger adjustments obtained with the Gaussian offsets. We also visualize this effect in \cref{fig:tightness}. We conduct additional ablations in \supp, and we also report some results for 3D pose estimation from sparse views on the Human3.6M dataset \cite{human36m}.

%% file: sections/conclusion.tex
\section{Conclusions and Discussion}\label{sec:con}
In this paper, we introduced \textit{\methodname}, a novel approach for human 3D representation that predicts 3D Gaussian Splatting \cite{gaussiansplatter}, enabling fast rendering with accurate poses. 
\methodname leverages multi-view rendering supervision to refine the 3D joint and body predictions. This dual capability combines precise pose estimation with novel view rendering, bridging two research paradigms and showcasing the benefits of our approach. Our method could be used to reduce the friction of training a pose estimation model to deploy to any novel sports setup, without requiring any expensive pose annotations or 3D scans.

\input{sections/limitation}

\minorsection{Acknowledgments}
This work was funded by the Podium Institute for Sports Medicine and Technology, at the University of Oxford. A.H. acknowledges the support of the KAUST Ibn Rushd Postdoc Fellowship program. J.H. acknowledges the support of the Royal Society (RG\textbackslash R1\textbackslash 241385), Toyota Motor Europe (TME), and EPSRC (VisualAI, EP/T028572/1).

%% file: sections/limitation.tex
\minorsection{Limitations}\label{sec:limit}
The main limitation in our method is the requirement of multi-view datasets to train.
Another issue is the slight blurriness that appears in some of the renderings, possibly as a result of the generalization limitations of the transformer when trained on relatively small datasets (in terms of subject diversity). A possible solution to this is to use a larger dataset or to pretrain on synthetic data. 


%% file: sections/supplement.tex
 \renewcommand{\thesection}{\Alph{section}}
\renewcommand{\thetable}{\Roman{table}}
\renewcommand{\thefigure}{\Roman{figure}}

\setcounter{section}{0}
\setcounter{table}{0}
\setcounter{figure}{0}
\newcommand{\parab}[1]{\smallskip\noindent\textbf{#1.}\,}

\section{Additional Results and Analysis} \label{secsup:results}
\subsection{Additional Ablations}
In addition to the ablations described in Table 5 in the main paper, we report here
three variations to the \methodname model that did not result in a performance improvement. The ablations are provided in Table \ref{tab:additional_ablation}. 

\parab{More Gaussians} The first design change we tested is an increase in the number of Gaussians per vertex. We increase the number of splats by predicting two or three independent offsets per vertex. Because random initialization breaks the symmetry, the model can learn to move each splat independently even though all two/three are anchored to the same vertex. Contrary to our assumption, an increase in the number of splats did not result in a increased visual quality of the renderings. 

\parab{Setting Opacity to 1} Predicting opacity is not strictly necessary to render humans, therefore we tried simplifying the model by removing this parameter. We removed the opacity prediction during training and manually set the opacity to 1 for all the Gaussians. 

\parab{Single-view + Multi-view Images} Next, to increase the subject diversity in the small datasets we use, we tried including some single view images in our training pipeline. For this experiment, we use crops of images containing humans from the MSCOCO dataset \cite{coco}. The single view images are used for training together with the multi-view images from the original dataset. For the single view images, the model predictions are supervised using the same input image. The results do not show any notable improvement.

\begin{table}[h]
\centering
\caption{\textbf{Additional Negative Ablations.} For completeness, we show additional ablations on HuMMan Dataset \cite{humman} that did not give positive improvements to our best setup of Table 5 in the main paper. For each setup, we report PSNR, SSIM, and LPIPS for novel view synthesis, as well as MPJPE (in mm) for 3D keypoints evaluation.  
}
\label{tab:additional_ablation}
\vspace{-0.5em}
\resizebox{0.99\linewidth}{!}{
    \centering
    \small{
    \addtolength{\tabcolsep}{-3.pt}
    \begin{tabular}{c cccc}
        \toprule
        Ablation setup & \multicolumn{3}{c}{Novel View} & \multicolumn{1}{c}{3D Shape} \\
         ~ & PSNR$\uparrow$ & SSIM$\uparrow$ & LPIPS$\downarrow$ & MPJPE (mm)$\downarrow$ \\
        \midrule
          our best model  & 21.79  & 0.87  & 0.12  &  50.8 \\
          2 Gaussians per vertex & 21.25 & 0.87 & 0.12 & 50.1 \\
          3 Gaussians per vertex & 21.18 & 0.87 & 0.12 & 53.2 \\
          setting opacity to 1 & 21.17 & 0.87 & 0.11 & 58.4 \\
        single-view + multi-view& 21.47 & 0.87 & 0.12 & 53.4 \\
        \bottomrule
    \end{tabular}}
}
\end{table}

\subsection{Overfitting Example}
\begin{figure*}[h]
    
    \begin{tikzpicture}[font=\footnotesize]

\node (A) at (0,0) {\includegraphics[width=0.9\textwidth]{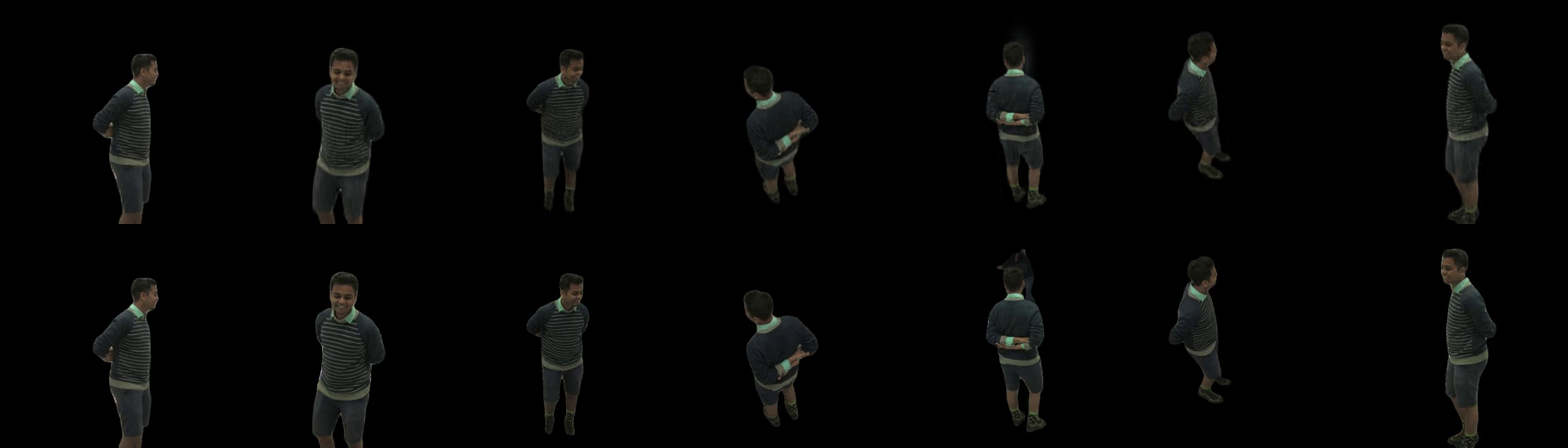}};
\node (B) [anchor=west] at ([yshift=1cm] A.east) {Renderings};
\node (C) [anchor=west] at ([yshift=-1cm]A.east) {Ground truth};

    \end{tikzpicture}
    \caption{\textbf{Overfitting to a single sample.} Ground truth (\textit{top}) and renderings (\textit{bottom}) of our model results when overfitting to a single data sample.}
    \label{fig:overfit}
\end{figure*}

To test that the number of Gaussians is sufficient to produce sharp details, we train our model to overfit a single data sample. We obtain an almost perfect reconstruction with PSNR of 41. Fig. \ref{fig:overfit} shows examples of the renderings we obtained. This result confirms our assumption that with a large enough dataset, our model would be able to learn sharper details than it currently learns on the small scale datasets.  

\subsection{Additional Details for TH21 Experiment}
\begin{figure*}[h]
    \begin{tikzpicture}[font=\footnotesize]

\node (A) at (0,0) {\includegraphics[width=0.415\textwidth]{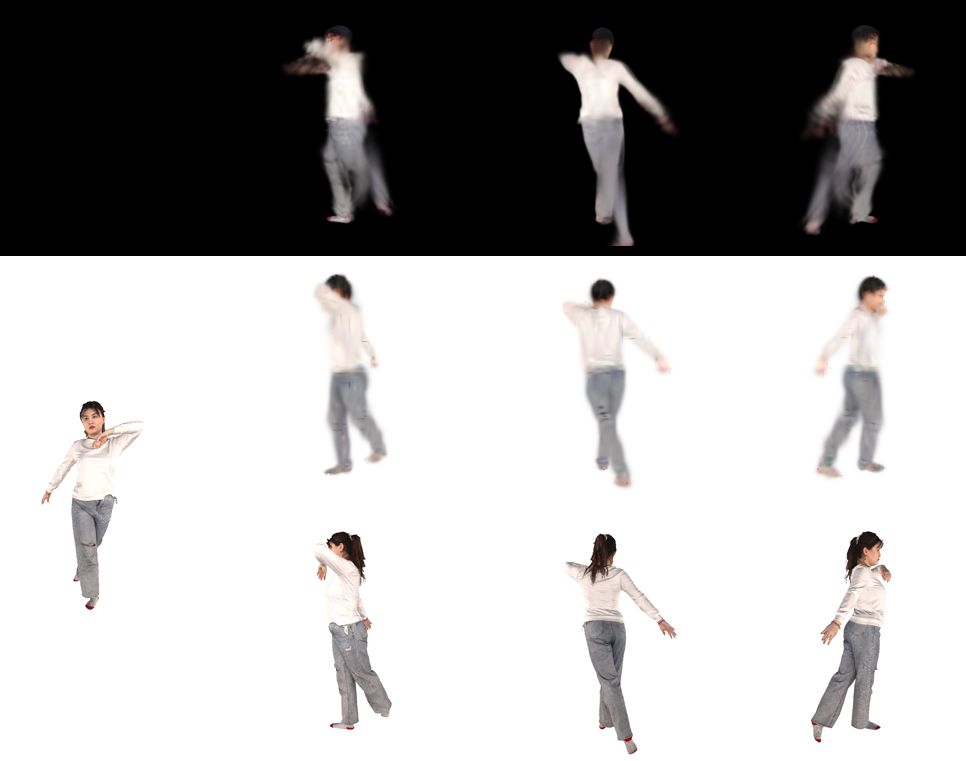}};
\node (B) [anchor=west] at ([yshift=-0.13cm] A.east) {\includegraphics[width=0.453\textwidth]{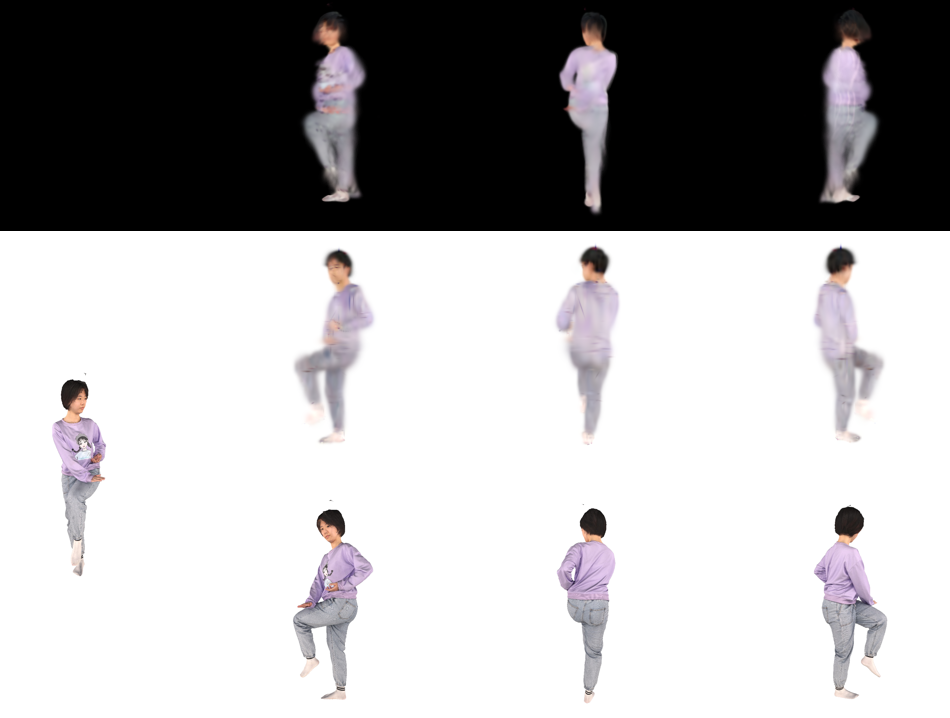}};
\node (C) [anchor=west] at ([yshift=1.5cm] B.east) {Splatter Image};
\node (D) [anchor=west] at (B.east) {\methodname (ours)};
\node (E) [anchor=west] at ([yshift=-1.5cm]B.east) {Ground truth};
    \end{tikzpicture}
    \caption{\textbf{Splatter Image comparison.} Side view comparison with Splatter Image \cite{splatter_image} on TH21 \cite{th2} for unusual input poses. Input image on the left, Splatter Image rendering in the first row, \methodname renderings in the second row.}
    \label{fig:splatterimage-comparison}
\end{figure*}

For the TH21 \cite{th2} experiment in Table 4 in the main report, we use 72 views rendered in a loop around the subject. We train both our method and Splatter Image \cite{splatter_image} using 256x256 images. Despite our model performing worse than Splatter Image in terms of visual metrics, our model also predicts the SMPL paramters for 3D pose estimation. This is both useful for downstream tasks, but also ensures that the underlying 3D shape is plausible for a human. This difference can be noticed in the examples in Fig. \ref{fig:splatterimage-comparison}, where \methodname can reconstruct a plausible human shape despite the uncommon input pose, while Splatter Image fails to reconstruct arms and legs. 

\subsection{3D Pose Estimation from Sparse Views}
We train \methodname on the common 3D pose estimation dataset Human3.6M \cite{human36m} using the default split for train and test subjects (subjects 9 and 11 are used for testing). This dataset is not ideal for our method as it only has 4 views and very few subjects, therefore it's difficult to generalize to unseen poses and subjects. Additionally, the human masks provided with the dataset are not always precise and our method tends to model parts of the background together with the human. This affects the visual results and the 3D pose estimation. The visual metrics for our \methodname are evaluated on a squared crop of size 256x256 around the human with a PSNR of 18.68 and a 3D error of MPJPE $\downarrow$ = 63.7 mm compared to 50.0 mm for HMR2 \cite{4dhumans}.

\begin{figure*}[t!]
  \begin{tikzpicture}[font=\footnotesize]
    \centering
\node (A) [anchor=west] at (0,0) {\includegraphics[width=0.95\textwidth]{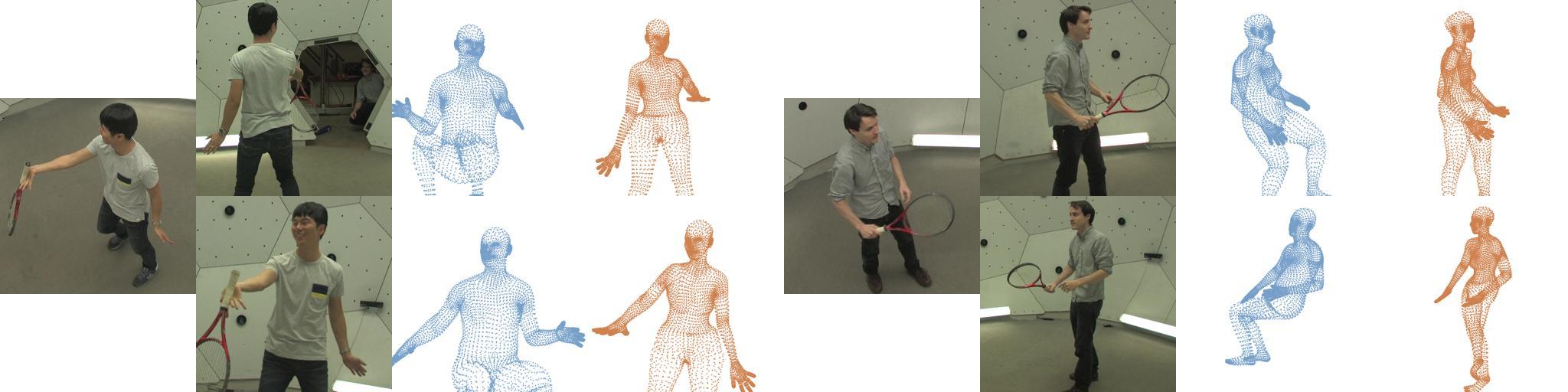}};
\node (B) [anchor=west] at ([xshift=-8.2cm, yshift=0.1cm] A.north) {Input Image};
\node (C) [anchor=west] at ([xshift=-6cm, yshift=0.1cm] A.north) {Other Views};
\node (D) [anchor=west] at ([xshift=-3.7cm, yshift=0.1cm] A.north) {HMR2 \cite{4dhumans}}; 
\node (E) [anchor=west] at ([xshift=-1.8cm, yshift=0.1cm] A.north) {\textbf{GST (ours)}};
\node (F) [anchor=west] at ([xshift=0.2cm, yshift=0.1cm] A.north) {Input Image};
\node (G) [anchor=west] at ([xshift=2.2cm, yshift=0.1cm] A.north) {Other Views};
\node (D) [anchor=west] at ([xshift=4.5cm, yshift=0.1cm] A.north) {HMR2 \cite{4dhumans}};
\node (E) [anchor=west] at ([xshift=6.5cm, yshift=0.1cm] A.north) {\textbf{GST (ours)}}; 
\end{tikzpicture}
\captionsetup{type=figure}
\captionof{figure}{\textbf{Example of human pose improvements using our method \methodname.} 3D human body results of our \methodname and SMPL predictions of HMR2 \cite{4dhumans} on a sports sequence from the CMU panoptic dome dataset \cite{cmu}.}
\label{fig:teaser_supp}
\end{figure*}

\section{Additional Visualizations} 
\begin{figure*}[t]
    \centering
\resizebox{0.9\hsize}{!}{
\setlength{\tabcolsep}{6mm}{
\begin{tabular}{ccccccc} 
 \multicolumn{1}{c}{Input Image} & \multicolumn{2}{c}{SHERF \cite{SHERF} w/ \cite{4dhumans} Renderings} & \multicolumn{2}{c}{\textbf{GST (ours)} Renderings} & \multicolumn{2}{c}{GT Renderings} \\ 
 \multicolumn{7}{c}{\includegraphics[width=0.99\textwidth]{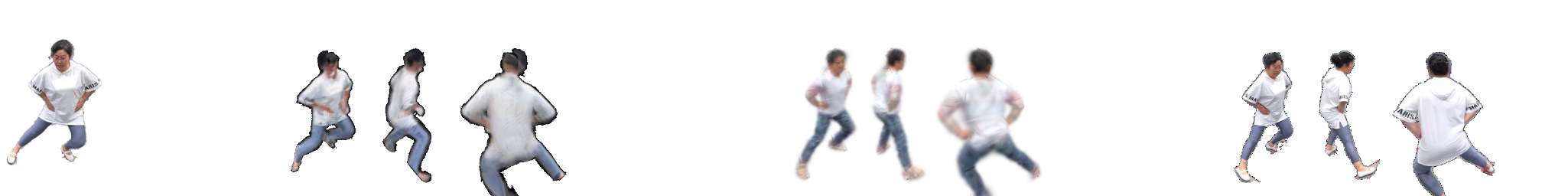}} \\ 
 \multicolumn{7}{c}{\includegraphics[width=0.99\textwidth]{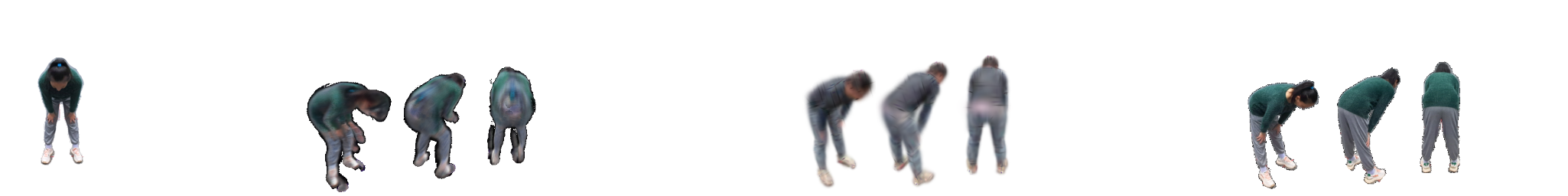}} \\ 
 \multicolumn{7}{c}{\includegraphics[width=0.99\textwidth]{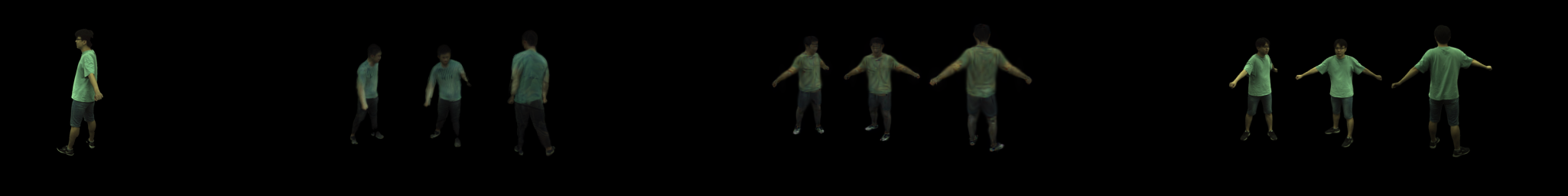}} \\ 
 \multicolumn{7}{c}{\includegraphics[width=0.99\textwidth]{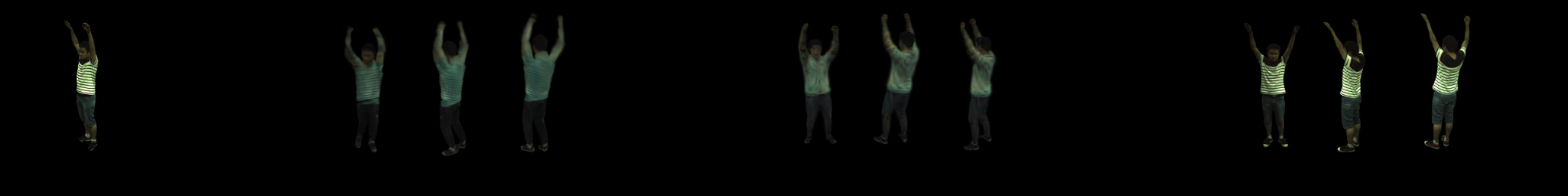}} \\

\end{tabular}
}
    }
       \caption{\small \textbf{Single Image NVS} on two subjects of Zju-Mocap \cite{ZJUMoCap} and two subjects of HuMMan \cite{humman} compared to SHERF \cite{SHERF} (after being adapted with HMR2 to work with single image input only). \methodname shows improved visual quality, especially when comparing the depicted pose to ground truth.
}
    \label{fig:sherf-comparison-appendix}
\end{figure*}

\begin{figure*}[h]
    \begin{tikzpicture}[font=\footnotesize]
    \centering
\node (A) [anchor=west] at (0,0) {\includegraphics[width=0.95 \textwidth]{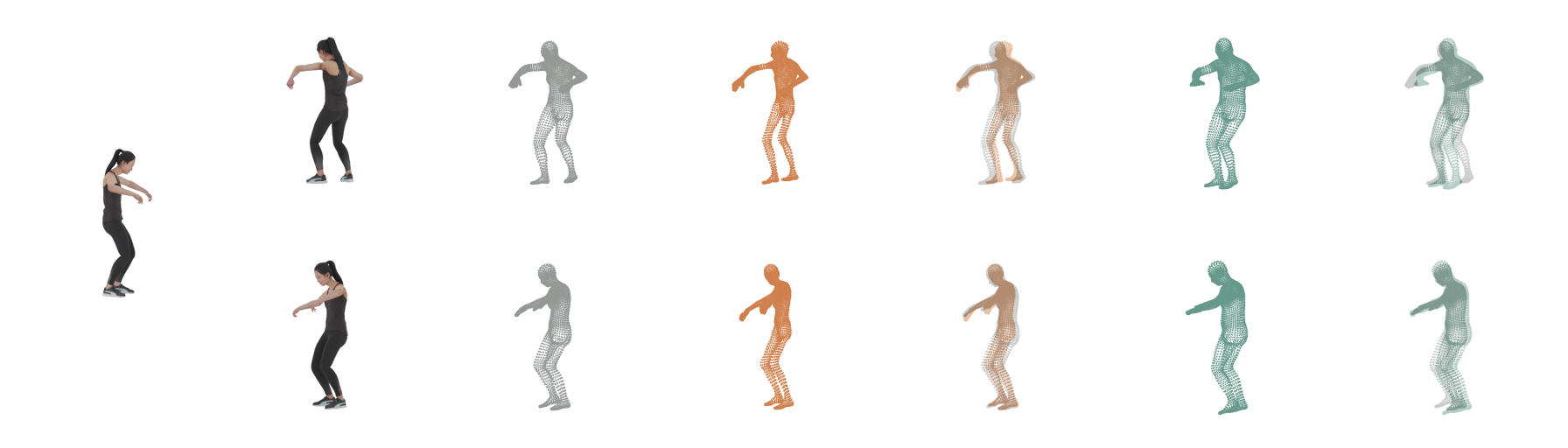}};
\node (D) [anchor=north] at (A.south) {\includegraphics[width=0.95 \textwidth]{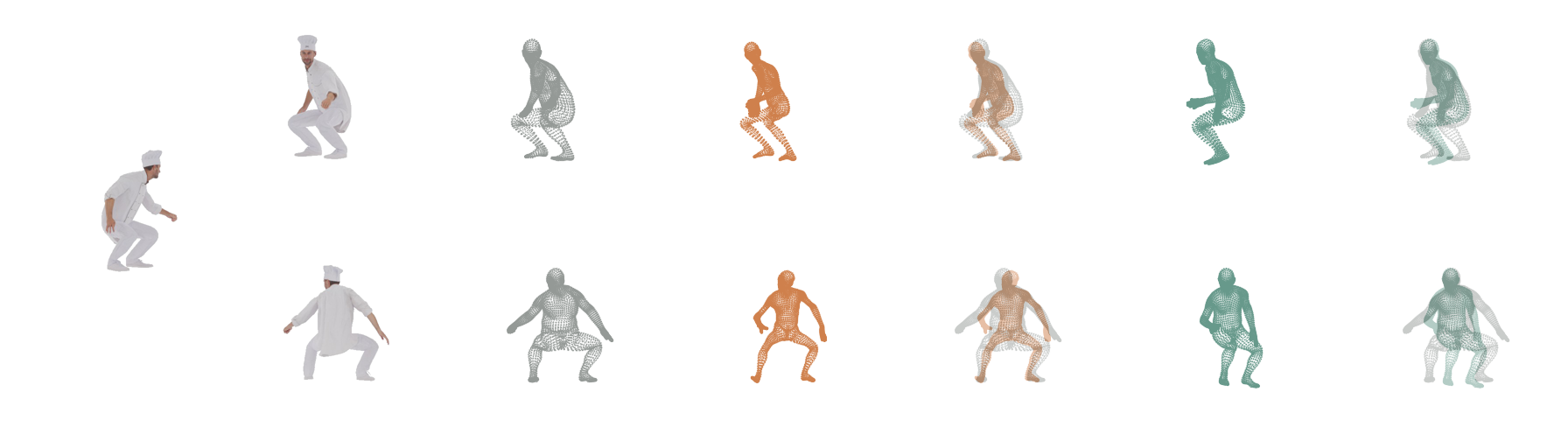}};
\node (B) [anchor=west] at ([xshift=-8cm,yshift=0.5cm] A.north) {Input Image};
\node (C) [anchor=west] at ([xshift=-5.5cm,yshift=0.5cm] A.north) {Other Views};
\node (L) [anchor=west] at ([xshift=-2.6cm,yshift=0.5cm] A.north) {GT};
\node (E) [anchor=west] at ([xshift=0.2cm,yshift=0.5cm] A.north) {\textbf{GST (ours)}};
\node (H) [anchor=west] at ([xshift=0.7cm,yshift=-0.1cm] E.south) {vs GT};
\node (F) [anchor=west] at ([xshift=5.1cm,yshift=0.5cm] A.north) {HMR2 \cite{4dhumans}};
\node (G) [anchor=west] at ([xshift=0.7cm,yshift=-0.1cm] F.south) {vs GT};
    \end{tikzpicture}
    \caption{\textbf{3D Shape Comparison with HMR2.} 3D human body results of our \methodname on two subjects of RenderPeople \cite{renderpeople} dataset compared to Ground Truth SMPL parameters \cite{SMPL}, and SMPL predictions of HMR2 \cite{4dhumans}.
}
    \label{fig:3dresults-appendix}
\end{figure*}

\begin{figure*}[h]
    \begin{tikzpicture}[font=\footnotesize]
    \centering
\node (A) [anchor=west] at (0,0) {\includegraphics[width=0.95 \textwidth]{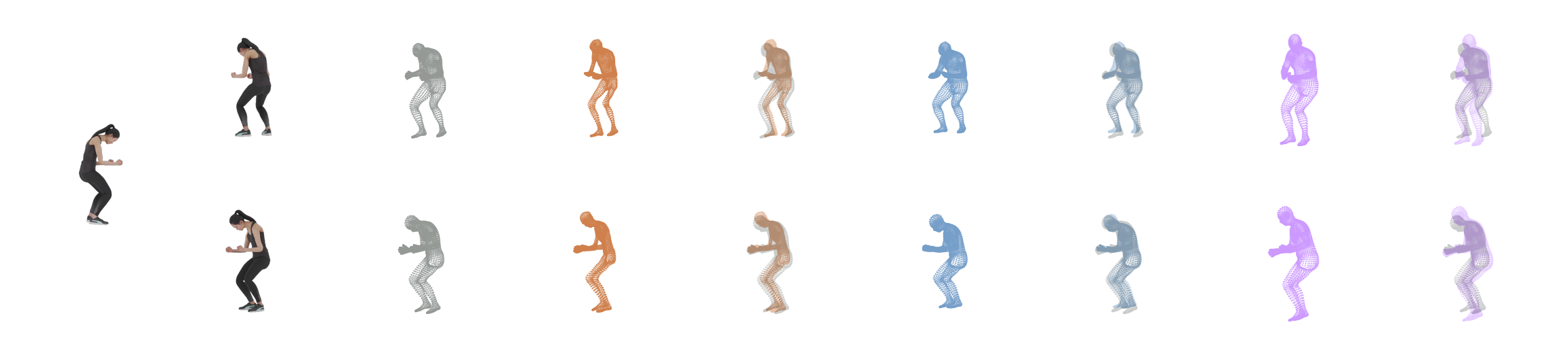}};
\node (D) [anchor=north] at (A.south) {\includegraphics[width=0.95 \textwidth]{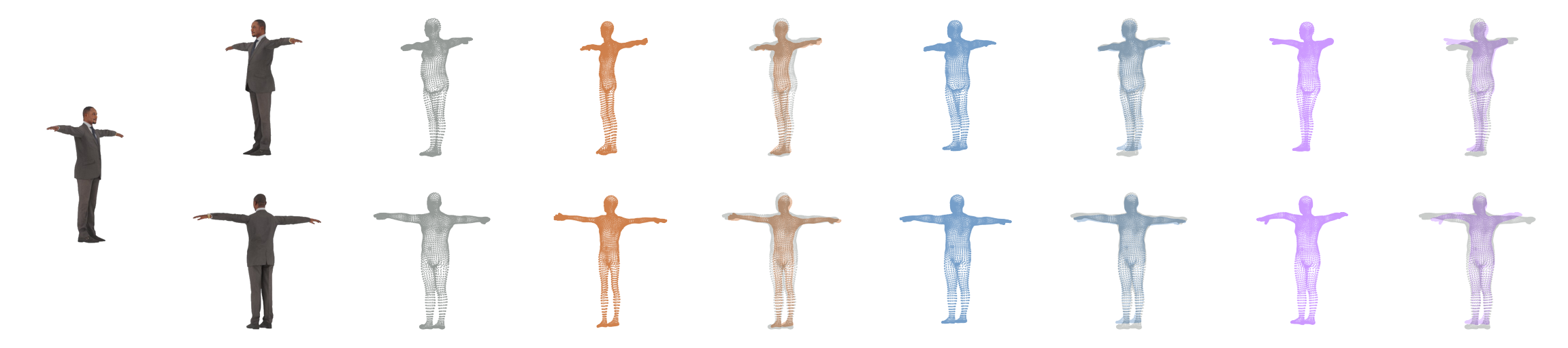}};
\node (W) [anchor=north] at (D.south) {\includegraphics[width=0.95 \textwidth]{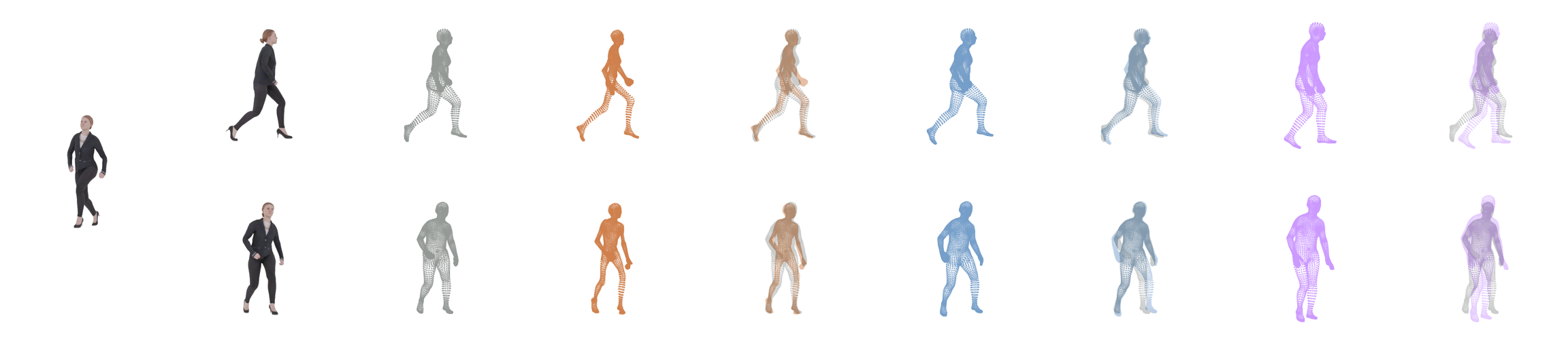}};
\node (X) [anchor=north] at (W.south) {\includegraphics[width=0.95 \textwidth]{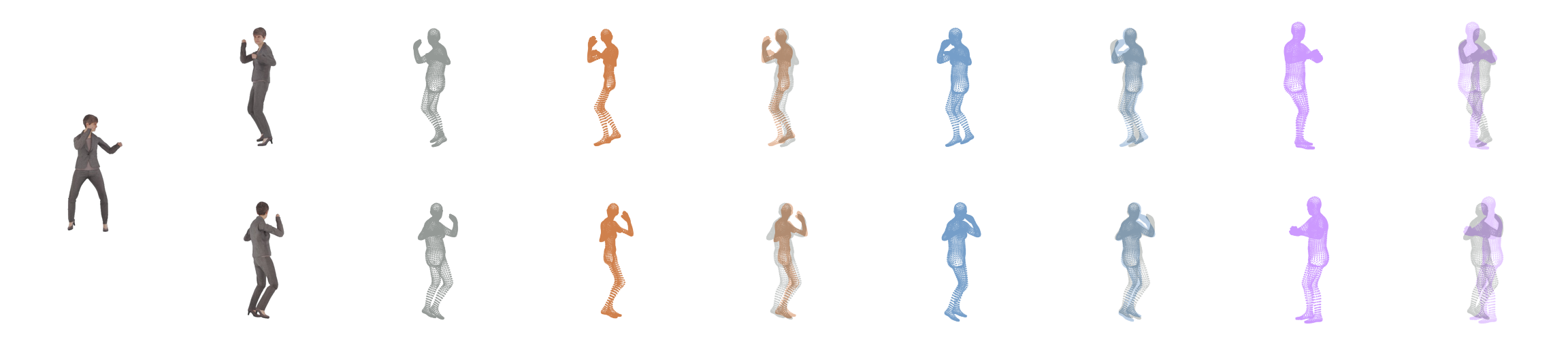}};
\node (Y) [anchor=north] at (X.south) {\includegraphics[width=0.95 \textwidth]{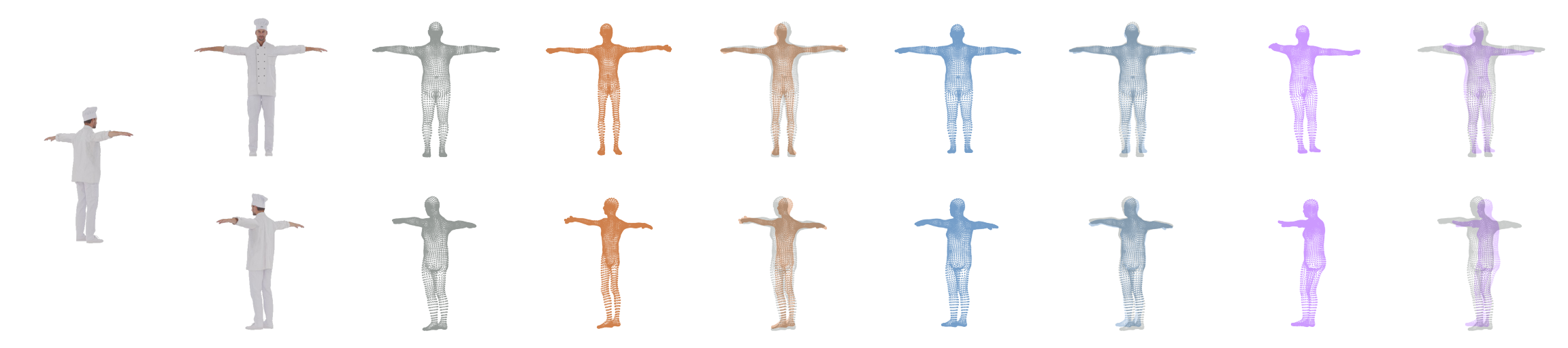}};
\node (B) [anchor=west] at ([xshift=-8.5cm,yshift=0.3cm] A.north) {Input Image};
\node (C) [anchor=west] at ([xshift=-6.2cm,yshift=0.3cm] A.north) {Other Views};
\node (Z) [anchor=west] at ([xshift=-4cm,yshift=0.3cm] A.north) {GT};
\node (E) [anchor=west] at ([xshift=-1.8cm,yshift=0.3cm] A.north) {\textbf{GST (ours)}};
\node (H) [anchor=west] at ([xshift=0.5cm, yshift=-0.1cm] E.south) {vs GT};
\node (F) [anchor=west] at ([xshift=1.4cm,yshift=0.3cm] A.north) {HMR2-3D finetuning};
\node (I) [anchor=west] at ([xshift=0.5cm, yshift=-0.1cm] F.south) {vs GT};
\node (G) [anchor=west] at ([xshift=5.1cm,yshift=0.3cm] A.north) {HMR2-2D finetuning};
\node (J) [anchor=west] at ([xshift=0.5cm, yshift=-0.1cm] G.south) {vs GT};
    \end{tikzpicture}
    \caption{\textbf{3D Shape Comparison with HMR2 After Fine-tuning on 2D and 3D Data.} 3D human body results of our \methodname on five subjects of RenderPeople \cite{renderpeople} dataset compared to Ground Truth SMPL parameters \cite{SMPL}, and SMPL predictions of HMR2 \cite{4dhumans}. We show two versions of HMR2, one finetuned on 2D data only (HMR2-2D), and one finetuned on 3D data (HMR2-3D). Our method is only finetuned on 2D image data, but the results are almost as accurate as HMR2 finetuned on 3D data. 
}
    \label{fig:3dresults_finetuned-appendix}
\end{figure*}

\begin{figure*}
\centering
\resizebox{0.67\hsize}{!}{
\setlength{\tabcolsep}{4mm}{
\begin{tabular}{ccccccc} 
 \multicolumn{7}{c}{\includegraphics[width=0.99\textwidth]{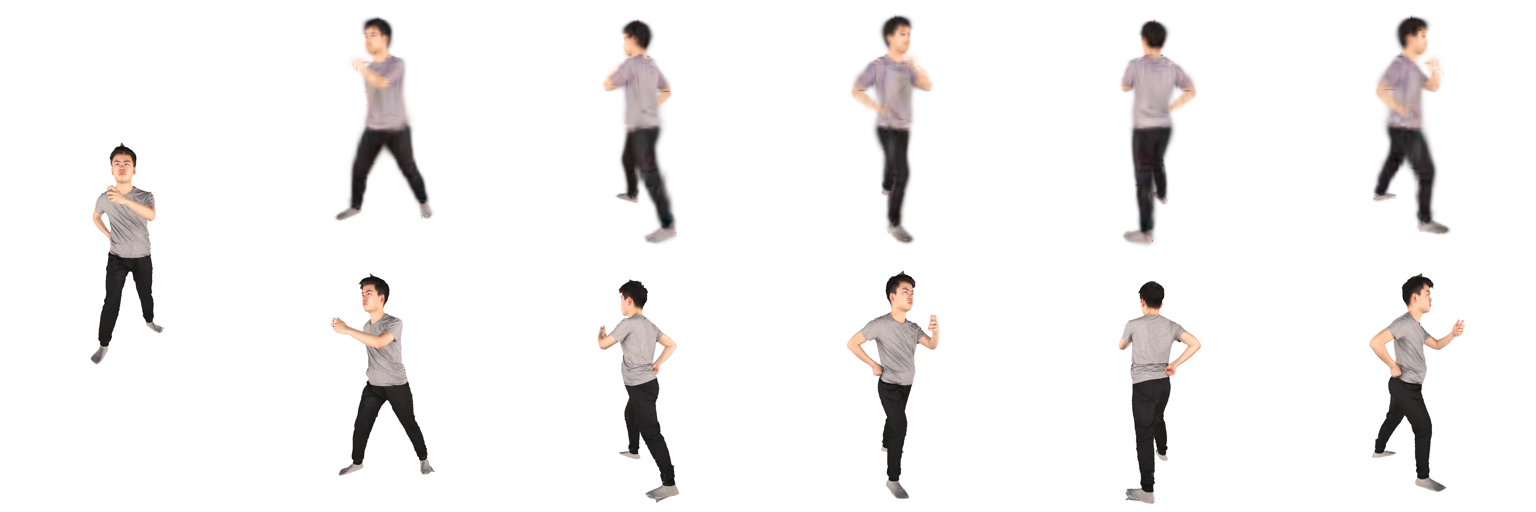}} \\ 
 \multicolumn{7}{c}{\includegraphics[width=0.99\textwidth]{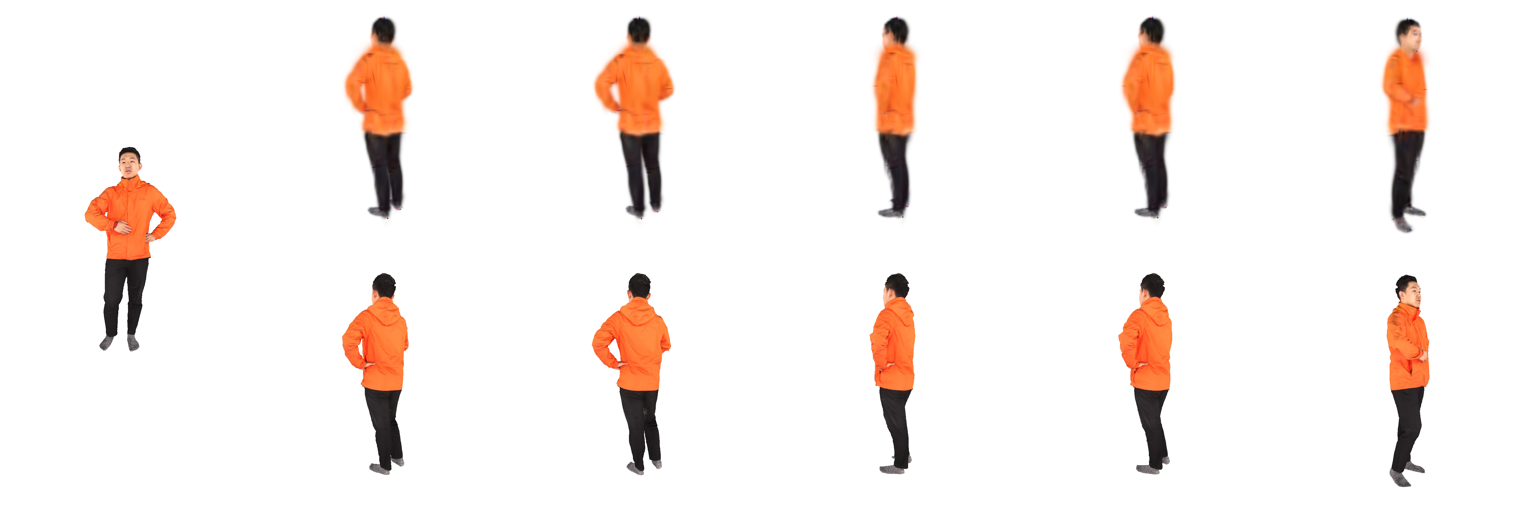}} \\ 
 \multicolumn{7}{c}{\includegraphics[width=0.99\textwidth]{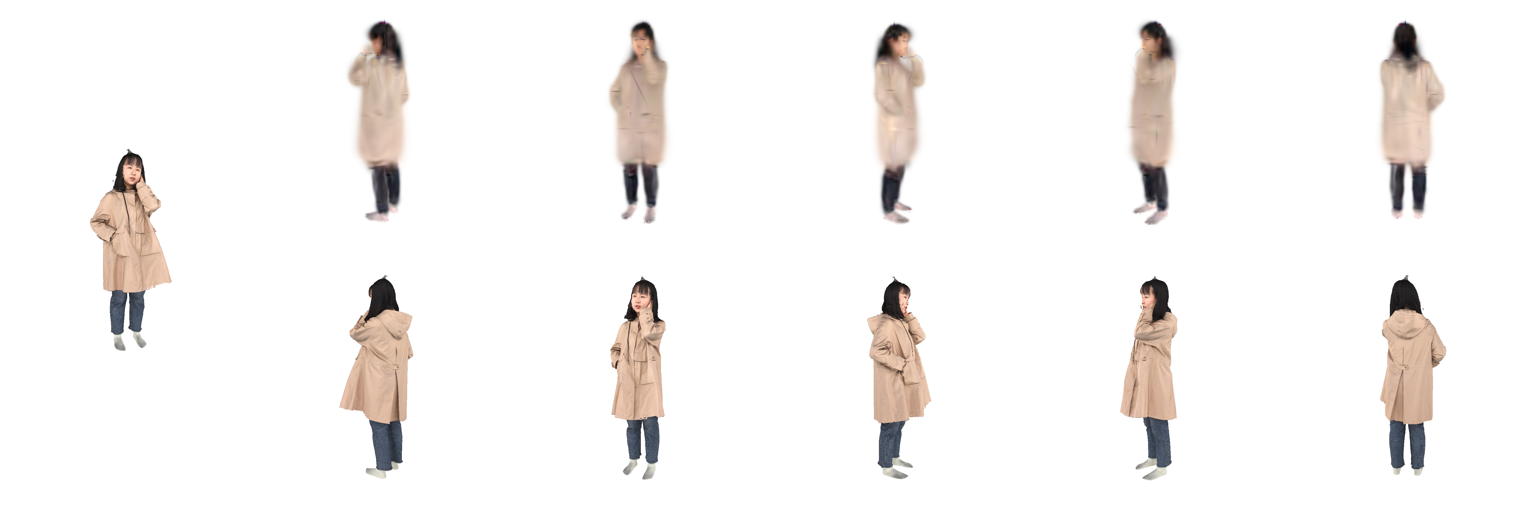}} \\ 
\end{tabular}
}
    }
\caption{\textbf{Results in TH21 \cite{th2}.} Rendering results for \methodname (\textit{top row}) compared to Ground Truth renderings (\textit{bottom row}) of each subject. An example of loose clothes is in the last row.} 
\label{fig:th21-appendix}
\end{figure*}

\begin{figure*}[t]
    \centering
\resizebox{0.97\hsize}{!}{
\setlength{\tabcolsep}{4mm}{
\begin{tabular}{ccccccc} 
 \multicolumn{7}{c}{\includegraphics[width=0.99\textwidth]{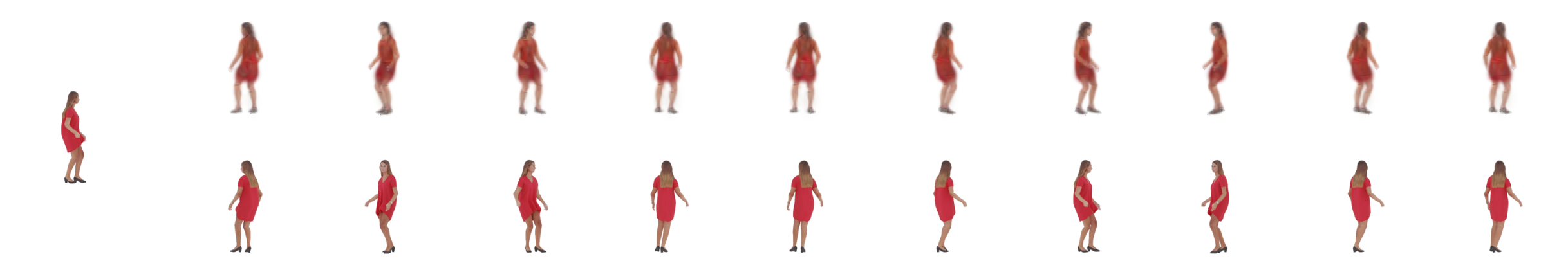}} \\ 
 \multicolumn{7}{c}{\includegraphics[width=0.99\textwidth]{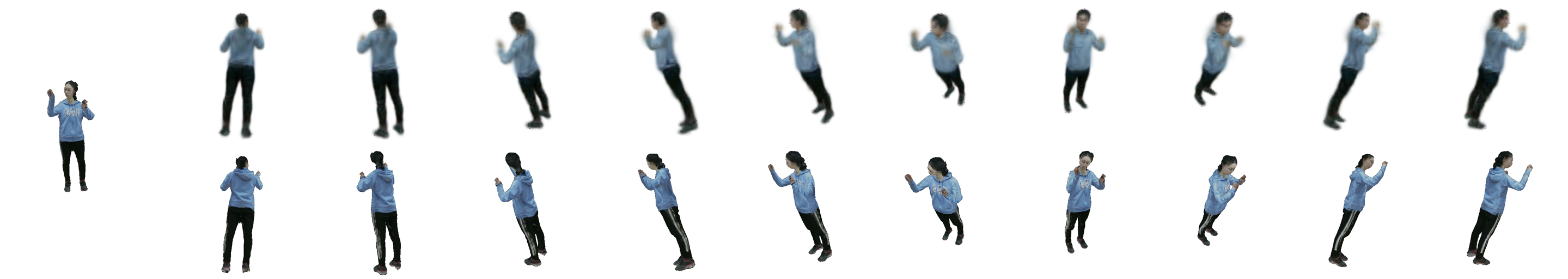}} \\ 

\end{tabular}
}
    }
       \caption{\small \textbf{Visualization of Single Image Novel View Synthesis Results on THuman and RenderPeople}. We show single image novel view synthesis results on one subject of THuman \cite{Thuman} dataset and one subjects of RenderPeople \cite{renderpeople} dataset of our \methodname (\textit{top row}) compared to Ground Truth renderings (\textit{bottom row}) of each subject.
}
    \label{fig:nvs_appendix_thuman_renderpeople}
\end{figure*}

Fig. \ref{fig:teaser_supp} shows additional pose estimation results on the sports sequence of the CMU Panoptic dataset \cite{cmu}. Fig. \ref{fig:sherf-comparison-appendix} shows additional examples of novel view synthesis comparisons with SHERF \cite{SHERF}. Fig. \ref{fig:3dresults-appendix} and \ref{fig:3dresults_finetuned-appendix} show additional pose comparisons for the RenderPeople \cite{renderpeople} dataset. Fig. \ref{fig:th21-appendix} and \ref{fig:nvs_appendix_thuman_renderpeople} show examples of novel view synthesis results for the TH21 \cite{th2}, THuman \cite{Thuman} and RenderPeople \cite{renderpeople} datasets. 

